\documentclass[11pt]{extarticle}
\usepackage[margin=1in]{geometry}
\usepackage{natbib}
\usepackage{setspace}
\usepackage{graphicx} 
\usepackage{algorithm}
\usepackage{algorithmic}
\usepackage{xcolor}         % colors
\usepackage{amsthm}
\usepackage{thmtools}
\usepackage{mathtools}
\usepackage{amsmath}
\usepackage{amssymb}
\usepackage{hyperref}
\usepackage{thmtools}
\usepackage{thm-restate}
\usepackage{url}

\newcommand\blfootnote[1]{
    \begingroup
    \renewcommand\thefootnote{}\footnote{#1}
    \addtocounter{footnote}{-1}
    \endgroup
}

\begin{document}

\title{FedBaF: Federated Learning Aggregation \\ Biased by a Foundation Model}

\author{Jong-Ik Park \and Srinivasa Pranav \and Jos\'e M. F. Moura \and Carlee Joe-Wong
}

\date{Electrical and Computer Engineering\\
Carnegie Mellon University \blfootnote{Authors are partially supported by NSF Grant CNS-2409138, CNS-2106891, CNS-2312761, and CCF-2327905. Srinivasa Pranav is partially supported by NSF Graduate Research Fellowships (DGE-1745016, DGE-2140739) and an ARCS Fellowship.}}

\maketitle

\begin{abstract}
Foundation models are now a major focus of leading technology organizations due to their ability to generalize across diverse tasks. Existing approaches for adapting foundation models to new applications often rely on Federated Learning (FL) and disclose the foundation model weights to clients when using it to initialize the global model. While these methods ensure client data privacy, they compromise model and information security. In this paper, we introduce Federated Learning Aggregation Biased by a Foundation Model (FedBaF), a novel method for dynamically integrating pre-trained foundation model weights during the FL aggregation phase. Unlike conventional methods, FedBaF preserves the confidentiality of the foundation model while still leveraging its power to train more accurate models, especially in non-IID and adversarial scenarios. Our comprehensive experiments use Pre-ResNet and foundation models like Vision Transformer to demonstrate that FedBaF not only matches, but often surpasses the test accuracy of traditional weight initialization methods by up to 11.4\% in IID and up to 15.8\% in non-IID settings. Additionally, FedBaF applied to a Transformer-based language model significantly reduced perplexity by up to 39.2\%.
\end{abstract}
\section{Introduction}
Developing foundation models~\citep{zhuang2024foundation} has become a major focus for leading technology companies like OpenAI, Microsoft, and Amazon AWS.
These deep learning models are often trained with vast amounts of high-quality data~\citep{bommasani2021opportunities} and their ability to generalize across different tasks and domains has made them essential assets for industry, government, and academia. Foundation models have been applied to natural language processing (e.g., text generation, translation, summarization), image generation and recognition, healthcare diagnostics, finance predictive analytics, and customer service and virtual assistant tasks~\citep{chen2023on, han2021pre, duan2021flexible, 10.1145/3533708, yosinski2014transferable}. 
When a foundation model's training data distribution overlaps with a new application, it provides a robust starting point for fine-tuning and customization.
Instead of training a model from scratch, with limited data and computes, we can leverage pre-trained foundation models to enable faster training.

For many applications, data that could be used to customize or fine-tune foundation models is often distributed across multiple clients, such as a network of clinics or small companies spread across different jurisdictions. For example, fine-tuning a recommendation model to fit a small company's product offering may require data from clients in various regions; and adapting a healthcare model for a network of clinics would involve confidential, distributed data sources.
Therefore, effective generalization requires access to diverse data from multiple clients~\citep{pranav2023peer}.

Federated Learning (FL) is a promising solution for fine-tuning these models without sharing client data: FL clients train models on diverse local data, and a central FL server aggregates the client updates to build and refine a global model~\citep{li2021survey,mcmahan2017communication,singh2019detailed,lyu2020threats,nguyen2021federated, FedMA_Wang2020Federated, FedLAMA_lee2023layer, siew2024fair}. Using a foundation model to initialize the global FL model leads to effective customization that leverages diverse, distributed data without directly accessing the client data~\citep{nguyen2022begin, chen2023on}. 
However, \textit{there are significant risks associated with sending a foundation model to clients, as required in traditional FL fine-tuning methods}.

\begin{figure}[h]
    \centering
    \includegraphics[width=\linewidth]{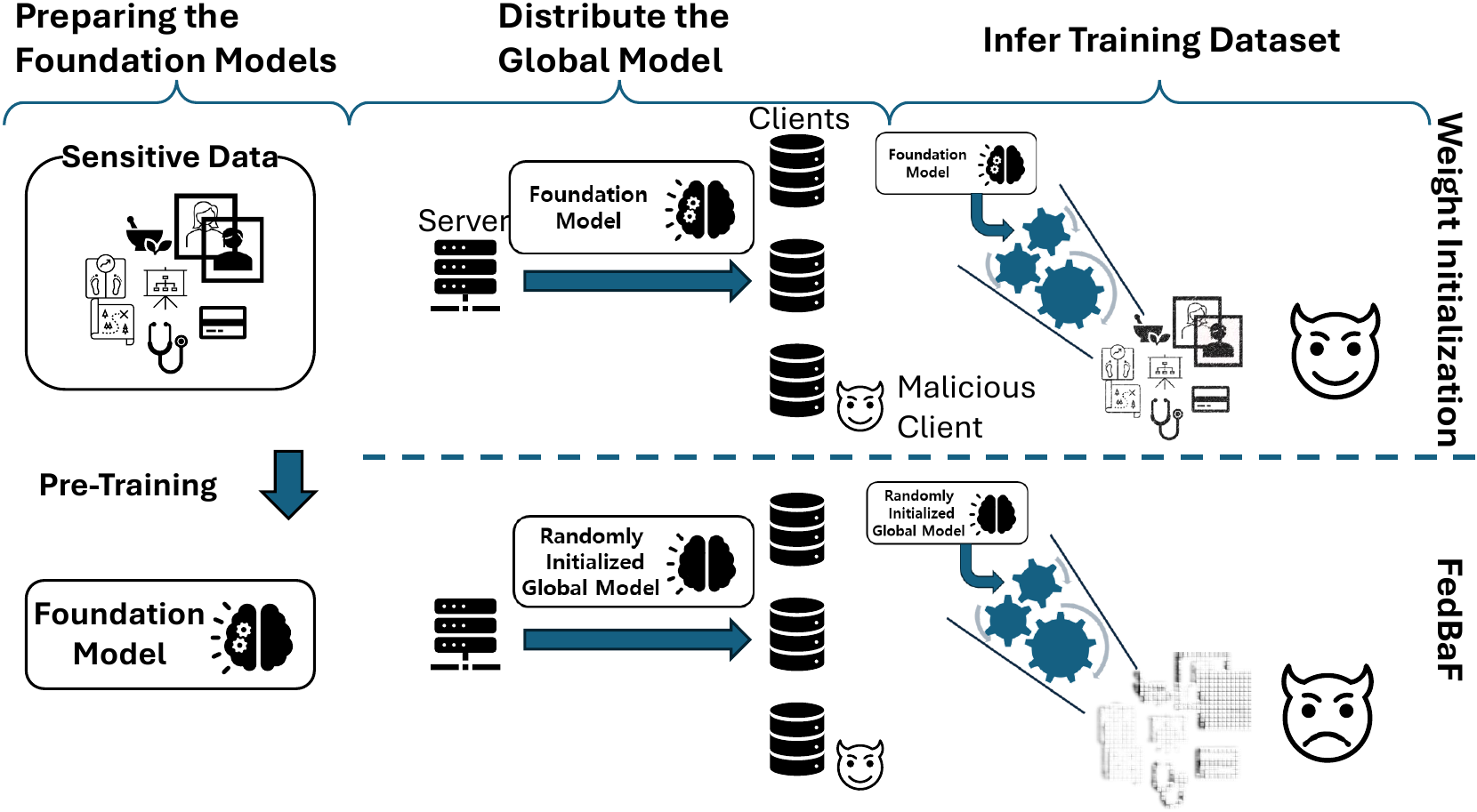}
    \vspace{-0.8 cm}
    \caption{A visualization of a Model Inversion Attack. Since FedBaF does not initialize the global model with a pre-trained foundation model, it becomes difficult for malicious clients to reconstruct the pre-training data from the distributed global model.}
    \label{fig:attack_desc}
\end{figure}

\textbf{First}, disclosing a foundation model's weights to FL clients poses a significant security risk. For example, malicious actors could carry out \textit{membership inference attacks}, identifying whether specific data was part of the foundation model’s training dataset. 
Then, an attacker could disrupt the global model's training by introducing updates that degrade performance on identified data through backdoor attacks, deliberately leading to targeted misclassification~\citep{dayal2023comparative, hu2022membership, wang2023gbmia}. 
Similarly, for \textit{model inversion attacks}, attackers use known model weights to reverse-engineer sensitive training data (see Figure~\ref{fig:attack_desc})~\citep{fredrikson2015model, li2022ressfl,zhang2020secret}. 
Protecting foundation models, often trained on sensitive, proprietary data, is critical for safeguarding the training data and maintaining model integrity~\citep{bagdasaryan2020backdoor, kim2023adversarial}.

\textbf{Second}, in competitive business contexts, disclosing foundation model weights to clients risks leaking strategic insights and proprietary information to adversaries~\citep{han2021pre, yu2023freedom}. This undermines a company's competitive advantage and substantial investments in data collection and training. 

\begin{figure}[h]
    \centering
    \includegraphics[width=\linewidth]{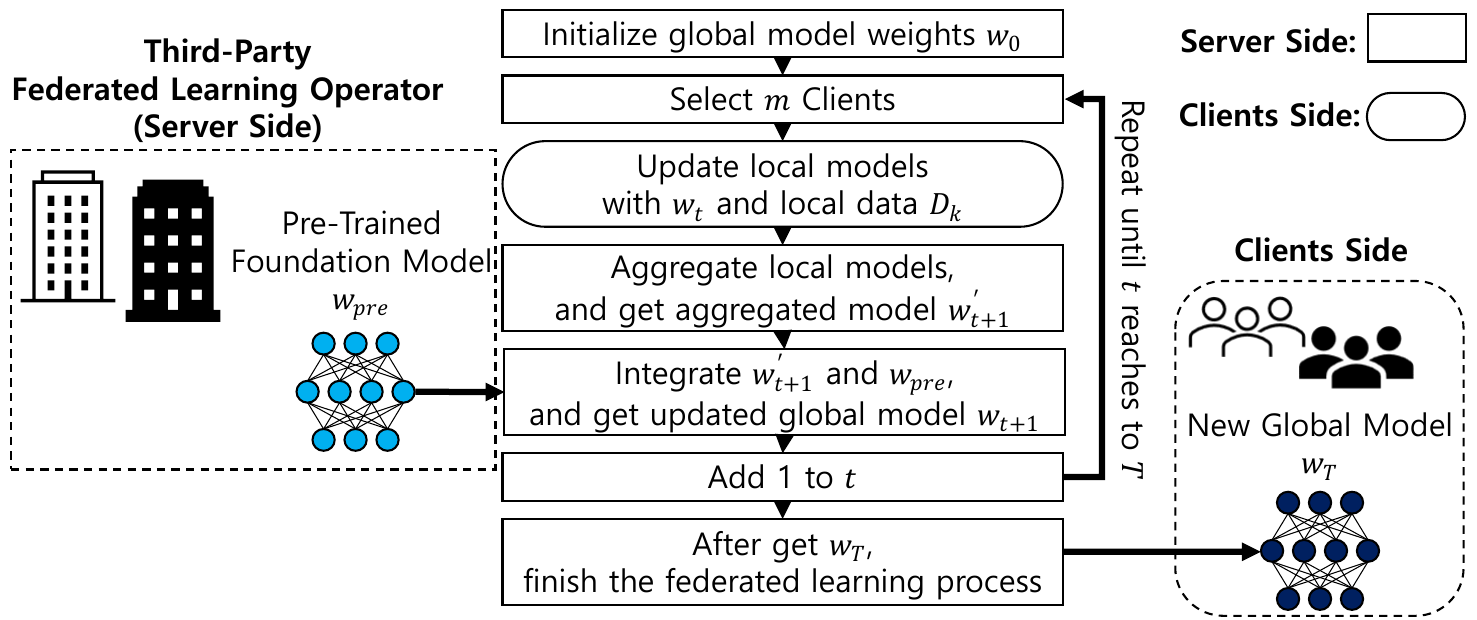}
    \vspace{-0.8 cm}
    \caption{Visualization of FedBaF: in each FL round's aggregation phase (after client updates), the server integrates a foundation model into the global model.}
    \label{fig:FLdiagram}
\end{figure}

To address these challenges, we present Federated Learning Aggregation Biased by a Foundation Model (\textbf{FedBaF}). Rather than using a foundation model to initialize the global model, FedBaF is a novel method for server-side foundation model integration during the task-specific global model aggregation phase of each FL round (see Figure~\ref{fig:FLdiagram}). Since the server uses the foundation model in the aggregation phase, FedBaF ensures that \textit{the foundation model is not disclosed to clients}.
FedBaF also gradually reduces the foundation model's influence as training progresses, thereby improving personalization for the client pool's data and matching or outperforming existing methods. 

FedBaF is particularly beneficial for a FL operator who owns the foundation model and needs to maintain security and integrity while fine-tuning it with a new set of clients. For instance, large technology companies such as Microsoft and Amazon, develop their own foundation models and often act as FL operators for domain-specific tasks across various industries.

When client data distributions are biased or non-IID, e.g., different ratios or a lack of certain labels, 
clients optimize correspondingly diverse local objective functions~\citep{pranav2023peerNonIID} and may send conflicting updates to the server that skew the global model~\citep{zhao2018federated}. In FedBaF, the foundation model continuously serves as a form of regularization and stabilizes the global model by reducing the influence of these conflicting updates during the aggregation phase~\citep{chen2023on, li2022domain, tan2022federated, yosinski2014transferable}.

Furthermore, FedBaF uses a fixed foundation model as an anchor and continuously incorporates it throughout the FL training process. This adds a layer of protection from adversarial attacks beyond those introduced by disclosing foundation model weights -- such as \textit{misclassification attacks} or \textit{backdoor attacks}, where compromised clients feed malicious updates to the server~\citep{lyu2020threats, bagdasaryan2020backdoor}.

\textbf{Our contributions:}\\
\textbf{1)} To the best of our knowledge, we are the first to propose an \textbf{algorithm that integrates foundation models into FL without distributing the foundation model to clients}. \\
\textbf{2)} We provide \textbf{theoretical analysis of FedBaF's effectiveness} that reveals how foundation models can promote convergence in non-IID (not independent and identically distributed) and non-convex settings. \\
\textbf{3)} We conduct extensive empirical evaluation and show that \textbf{FedBaF matches or exceeds the training performance of traditional weight initialization methods} -- with better test performance in 10 out of 14 cases. Our experiments use
Pre-ResNet and more complex architectures like Vision Transformer and Transformer-based language models frequently used as foundation models~\citep{xu2023multimodal,kenneweg2024foundation}.
Compared to standard FedAvg~\citep{mcmahan2017communication} and FedProx~\citep{li2020federated} with foundation models used for weight initialization, FedBaF achieves accuracy improvements of up to 10.8\% in IID and up to 37.5\% globally and 5.9\% locally in non-IID settings. Simultaneously, FedBaF safeguards the foundation model.
Similarly, applying FedBaF to a Transformer-based language model significantly reduced perplexity by up to 76.0\%.

Under adversarial misclassification attacks, FedBaF demonstrates increased robustness by improving FedAvg and FedProx test performance by up to 19.4\% in IID environments, up to 64.7\% globally, and 7.2\% locally in non-IID environments. Additionally, in 8 out of 12 cases, FedBaF was more robust than traditional weight initialization methods.

We outline related works in Sec.~\ref{sec: related works}. We then detail our approach, FedBaF, in Sec.~\ref{sec: methodology}. In Sec.~\ref{sec: analysis}, we present theoretical analysis and, in Sec.~\ref{sec: experiments}, we provide extensive experimental evaluation. Finally, we conclude our research findings and discussion in Sec.~\ref{sec: conclusion}.

\section{Related Work} \label{sec: related works}
Traditionally, pre-trained models are used in FL to initialize the weights of the global FL model. The server distributes this model to local clients, and the clients update it by using their local data. We refer to this approach as ``weight initialization'' throughout this paper. Such fine-tuning of a pre-trained model can significantly improve performance of the learned FL global model by integrating data from new clients~\citep{nguyen2022begin}. Several recent studies devised methods that leverage weight initialization to further improve performance: Federated Nearest Class Means (FedNCM)~\citep{guiding_last_layer_NEURIPS2023_dcc0ac74} for last-layer guidance, Federated Recursive Ridge Regression (Fed3R)~\citep{fed3R_pmlr-v235-fani-24a}, Fractal Pair Similarity (FPS)~\citep{chen2023on}, and  FedPCL~\citep{tan2022federated}. 

Several works also explore the use of foundation models in FL. These include cases where a subset of the weights of a large foundation model are chosen to initialize and fine-tune a smaller model~\citep{larger_ones_xu2024initializing} and when clients have diverse model architectures~\citep{wang2023flexifed, park2024federated}.
Particularly in scenarios with limited pre-training data~\citep{chen2023on}, approaches often rely on synthetic data~\citep{nikolenko2021synthetic, chen2023on} for pre-training.
Federated Prototype-wise Contrastive Learning (FedPCL) is a significant development that improves communication efficiency in FL by using class prototypes~\citep{tan2022federated} and enhances personalized learning by having clients share class-specific information more effectively. 

The related works discussed in this section so far achieve good performance, but they do not consider the significant \textbf{security vulnerabilities} that result from sharing a foundation model with local clients, which compromise data privacy and the integrity of the global model. Malicious clients with access to foundation models can exploit them through: Model Inversion Attacks, recovering original training data or sensitive attributes from the model's outputs~\citep{fredrikson2015model, zhang2020secret, li2022ressfl}; Membership Inference Attacks, analyzing model predictions to determine whether specific data records were used in training~\citep{hu2022membership, dayal2023comparative, wang2023gbmia}. 
These attacks compromise the security of the FL system, necessitating the development of more secure methods for leveraging pre-trained foundation models in FL settings.

FedBaF addresses these security challenges by not sharing the foundation model with clients during the weight initialization stage. Instead, it dynamically integrates the foundation model's pre-trained weights during the aggregation phase of each training round. We show that FedBaF improves privacy while matching or exceeding the performance achieved by weight initialization.

\section{Methodology} \label{sec: methodology}
In this section, we introduce FedBaF, whose approach is illustrated in Figure~\ref{fig:FLdiagram}. FedBaF involves the server repeatedly leveraging pre-trained foundation model weights throughout the aggregation phases of the FL training process. 
For example, the pre-trained weights corresponding to feature extraction layers provide valuable representation mappings that guide the new model's feature extractor during training.

To mimic the performance gains of weight initialization, the server uses the foundation model as a strong anchor in the earlier FL rounds. To enable the FL global model to evolve and fit the clients' data as FL training continues, the server assigns rapidly decaying importance to the foundation model that is on the order of $1/\sqrt{t}$ and proportional to the change in model parameters caused by client updates. To further maintain foundation model confidentiality, the server randomly samples the aggregation weights (or importance) of the foundation model from a uniform distribution in each round. These aspects of FedBaF maintain foundation model privacy while enabling further application-specific tuning to achieve performance on par or exceeding that of weight initialization methods. 

\begin{algorithm}[ht]
\caption{Federated Learning Aggregation Biased by a Foundation Model (FedBaF).
}
\begin{algorithmic}[1]
\STATE Initialize global model weights $\mathbf{w}_0$
\FOR{each round $t = 0, 1, 2, \ldots, T$}
    \STATE $m \gets \max(C \cdot K, 1)$ \\
    \STATE $S_t \gets$ (random set of $m$ clients)
    \FOR{each client $k \in S_t$ \textbf{in parallel}}
        \STATE $\mathbf{w}_{t+1}^k \gets$ ClientUpdate($\mathbf{w}_t, D_k$)
    \ENDFOR
    \STATE $\mathbf{w}'_{t+1} \gets \sum_{k \in S_t} \frac{n_k}{\sum_{\ell \in S_t} n_\ell} \mathbf{w}_{t+1}^k$
    \STATE $\tau_t \gets \frac{\left\|\frac{\mathbf{w}'_{t+1}}{\|\mathbf{w}'_{t+1}\|} - \frac{\mathbf{w}_{t}}{\left\|\mathbf{w}_{t}\right\|}\right\|}{\sqrt{t+1}}$
    \STATE $\alpha_t \gets \frac{\psi}{\tau_0}\text{U}(1,2)$ 
    \STATE $\mathbf{w}_{t+1} \gets \frac{1}{1 +\alpha_t\tau_t} (\mathbf{w}'_{t+1} + \alpha_t\tau_t \left(\mathbf{w}_{pre} \setminus \mathbf{w}_t )\right)$
\ENDFOR \\
\STATE \textbf{ClientUpdate}$(\mathbf{w}, D)$ 
    \STATE \quad Initialize local model weights with $\mathbf{w}$
    \STATE \quad Update local model weights using local data $D$
    \STATE \textbf{return} updated model weights \\
\end{algorithmic}
\label{algorithm: adapting pre-trained model}
\end{algorithm}
Algorithm~\ref{algorithm: adapting pre-trained model} describes how FedBaF fits into the traditional FL framework by incorporating foundation model weights during aggregation, as illustrated in \textit{Lines 9-11}~\citep{mcmahan2017communication}. FedBaF is versatile and can be embedded into many existing FL algorithms, e.g., SCAFFOLD~\citep{karimireddy2020scaffold}, FedProx~\citep{li2020federated}, FedAdam~\citep{reddiadaptive}, or other FL strategies, by modifying their aggregation methods (\textit{Lines 8-11}) and using their existing \textbf{ClientUpdate}$(\mathbf{w}, D)$ logic for clients' local training in \textit{Line 13}.

\textbf{Modifying FL aggregation.} FedBaF's aggregation process in each training round begins with the aggregation step of an existing FL algorithm, which, as mentioned above, includes FedAvg, SCAFFOLD, FedProx, and FedAdam. To illustrate an example with FedAvg, \textit{Line 8} of Alg.~\ref{algorithm: adapting pre-trained model} uses FedAvg's aggregation step and computes a weighted sum of the updated model parameters from each client. After this aggregation, \textit{Line 11} incorporates the pre-trained model weights ($\mathbf{w}_{pre}$) into the global FL model, controlled by the factor $\tau_t$ defined in \textit{Line 9}.

Here, $(\mathbf{w}_{pre} \setminus \mathbf{w}_t)$ refers to \textit{the subset of layers from the foundation model} ($\mathbf{w}_{pre}$) that have the same architecture as the corresponding layers in the global FL model ($\mathbf{w}_t$), ensuring that only compatible layers are used during aggregation. When the foundation and FL models have the same architecture except that the input and output layers (i.e., first and last layers) differ due to variations in input features or the number of classes, FedBaF excludes these input and output layers from the aggregation and integrates only the shared intermediate (hidden) layers. The differing layers are randomly initialized and then trained using data from the client pool, as in standard FL.

\textbf{Foundation and Global Model Architecture Mismatch.}  
In practice, the foundation model and the global model architectures may differ beyond the input and output layers. These differences can arise in terms of the number of layers or the number of parameters per layer. Since foundation models are typically larger than global FL models -- consistent with their role as highly expressive networks pre-trained on extensive data~\citep{meng2023foundation, awais2025foundation} -- we consider the following two cases: 

$\bullet$ \textbf{Foundation model with more layers:} When the foundation model is deeper than the global FL model, only a subset of its layers is used during aggregation. We take advantage of the fact that most large models have layers grouped into \textit{sections}. Here, a section is a contiguous subset of layers within a model that shares similar structural properties, such as the number of parameters, functional roles, or connectivity patterns. FedBaF selects and matches sections between the foundation and global models, prioritizing feature extraction sections near the input layer.
Within each matched section, only layers that align with the global model's architecture are integrated into FL training. 
Methods for selecting compatible layers during aggregation are explored in~\cite{park2024federated, larger_ones_xu2024initializing}.

$\bullet$  \textbf{Foundation model with more parameters per layer:} If the foundation model has layers with more parameters than the global model, only a subset of parameters within each section and layer is selected to match the global FL model. Since different sections may have varying parameter distributions, the aggregation is performed iteratively per section and per layer.
To consider these mismatches, \textit{Line 11} of Alg.~\ref{algorithm: adapting pre-trained model} is expanded:

\begin{algorithmic}[1]
    \FOR{each section $s$ in shared sections between $\mathbf{w}_{pre}$ and $\mathbf{w'}_t$}
        \FOR{each layer $l$ in section $s$}
            \FOR{each parameter subset $p$ in layer $l$ up to $P_t^{(s,l)}$ parameters}
                \STATE $\mathbf{w}_{t+1}^{(s,l,p)} \gets \frac{1}{1 +\alpha_t\tau_t} \left(\mathbf{w'}_{t+1}^{(s,l,p)} + \alpha_t\tau_t \mathbf{w}_{pre}^{(s,l,p)}\right)$
            \ENDFOR
        \ENDFOR
    \ENDFOR
\end{algorithmic}
Here, $P_t^{(s,l)}$ denotes the number of parameters selected per layer $l$ within section $s$. Only the first $P_t^{(s,l)}$ parameters in each layer are incorporated into the global model.

FedBaF can be analogously modified to handle the case where the foundation model is smaller than the global FL model. By applying these modifications, FedBaF can seamlessly adapt to different network architectures, ensuring that foundation model knowledge is effectively transferred while maintaining structural compatibility with the FL model.

\textbf{Designing $\tau_t$.} Our careful design of $\tau_t$ uses the $L2$ norm of the difference between consecutive normalized weights of $\mathbf{w'}_{t+1}$ and $\mathbf{w}_{t}$, divided by $\sqrt{t+1}$.  
This change in the model's weights between rounds reflects how much the global model adapts to new client updates.
The normalization prevents $\tau_t$ from becoming too large.  The factor $\sqrt{t+1}$ ensures that, as training progresses, the influence of $\mathbf{w}_{pre}$ gradually diminishes, but not too quickly, and $\mathbf{w}_{t+1}$ approaches the improving averaged weights $\mathbf{w'}_{t+1}$. This strategy is critical to keeping the global model flexible and effective, especially when client data differs from the data used to train the foundation model~\citep{karimireddy2020scaffold}.

Depending on the network architectures (e.g., the number of weights or scale of the initialized weights), the scale of $\tau_t$ can vary. In non-IID situations and during adversarial attacks, the factor $\tau_t$ becomes significant. In particular, a large $\tau_t$ can indicate the presence of non-IID data or an attack, as such scenarios often result in large differences in consecutive weight updates.
To keep $\tau_t$ within a suitable range, we introduce the parameter $\alpha_t$ in \textit{Line 10}, which depends on $\tau_0$ and the hyper-parameter $\psi$. We empirically find that setting $\alpha_t$ such that $\alpha_t \tau_0$ is less than 2 in the initial round ($t=0$) prevents excessively large values that could overly bias the global model towards the foundation model. A lower bound of 1 for $\alpha_t \tau_0$ also ensures that the minimum impact of the foundation model is significant for small $t$. We thus ensure the influence of the foundation model in the critical initial training stages, while still allowing the global model to adapt as training progresses.

\textbf{Designing $\alpha_t$.} Sampling $\alpha_t$ from the uniform distribution $\frac{\psi}{\tau_0}\text{U}(1,2)$ for every round makes it difficult for clients to reverse-engineer the foundation model, thus meeting FedBaF's model security guarantees. To see this, \textit{Line 11} can be rearranged for \(\mathbf{w}_{pre}\) as 
\begin{equation*}
    \mathbf{w}_{pre} = \frac{(1+\alpha_t\tau_t)\mathbf{w}_{t+1} - \mathbf{w'}_{t+1}}{\alpha_t\tau_t}.
\end{equation*}
In the worst-case scenario, where all local clients are malicious and collaborating to extract the foundation model's weights, they can access \(\tau_t\), \(\mathbf{w}_{t+1}\), and \(\mathbf{w'}_{t+1}\). However, because \(\alpha_t\) is randomly chosen in each round $t$ and known only to the server, the foundation model's weights cannot be extracted. If \(\alpha_t\) were static, even if the server did not disclose it, malicious clients could determine this constant value by solving the residual equations from two successive rounds,
\begin{equation*}\frac{(1+\alpha_t\tau_t)\mathbf{w}_{t+1} - \mathbf{w'}_{t+1}}{\alpha_t\tau_t} = \frac{(1+\alpha_{t+1}\tau_{t+1})\mathbf{w}_{t+2} - \mathbf{w'}_{t+2}}{\alpha_{t+1}\tau_{t+1}}. 
\end{equation*}
This would eventually reveal the foundation model's weights. \textit{More details, along with empirical analysis on the role of \(\alpha_t\) and the security advantages of FedBaF are provided in Appendix~\ref{appendix:security}.}

We formally examine this idea in the next section.

\section{THEORETICAL ANALYSIS} \label{sec: analysis}
In this section, we focus on deriving performance guarantees for FedBaF, focusing on its convergence properties.
Sec.~\ref{subsec:convergence} examines the general convergence behavior of FedBaF, while Sec.~\ref{subsec:nonIID} shows specifically on how FedBaF manages convergence in the presence of diverse, non-IID local client data distributions.

The following notation, problem setup, and assumptions are used throughout our analysis. Given $m$ clients, let the $k$th device's training data be drawn from $\mathcal{D}_k$. The FL problem can be formulated as the following global objective,
\begin{equation}
\label{eq:global_objective}
\min_{\mathbf{w}} \frac{1}{\sum_{k=1}^{m} n_k} \sum\limits_{k=1}^{m} n_k f_{\mathcal{D}_k}(\mathbf{w}),
\end{equation}
where $\mathbf{w}$ are model (usually, deep neural network) weights and the $f_{\mathcal{D}_k}$ are $L$-smooth local objective functions.
The convergence analysis presented in this section also makes the following standard assumptions made by \citep{karimireddy2020scaffold} and detailed in Appendix~\ref{appendix:theory}. Each client locally optimizes $\mathbf{w}$ using stochastic gradient descent, where the stochastic gradients are (i) unbiased and (ii) have bounded variance.
We also assume (iii) bounded gradient dissimilarity: the norm of the difference between the gradient of the global objective and the gradients computed using different local objective functions is bounded.  Lastly, we assume that (iv) the foundation model has the same architecture as the global model, ensuring compatibility during aggregation. See Appendix~\ref{sec:appendix_assumptions} for mathematical details regarding the assumptions.

\subsection{General Convergence Analysis}\label{subsec:convergence}

\begin{restatable}{proposition}{improvementprop}
\label{prop:improvement}
    Let $\mathbf{w}^*$ be a (bounded) local minimum of the global objective function in \eqref{eq:global_objective}. Consider an FL algorithm that converges to $\mathbf{w}^*$
    and let $\mathbf{w}'_{t}$ be its global model in each training round $t$. Suppose we run the same algorithm but using FedBaF for the aggregation, and let $\mathbf{w}_{t}$ be the FedBaF global model at round $t$. Let $\alpha_t$ satisfy 
    \begin{equation}
    \label{eq:alpha_condition}
        \alpha_t < \frac{2\|\mathbf{w}'_{t+1} -\mathbf{w}^*\|^2}{(\|\mathbf{w}_{\text{pre}} -\mathbf{w}^*\|^2-\|\mathbf{w}'_{t+1} -\mathbf{w}^*\|^2)\tau_t}
    \end{equation}
    for all $t$ where $\|\mathbf{w}'_{t+1} -\mathbf{w}^*\|^2 < \|\mathbf{w}_{\text{pre}} -\mathbf{w}^*\|^2$. Then $\forall t\;\|\mathbf{w}_{t} - \mathbf{w}^*\| < \|\mathbf{w}'_{t} - \mathbf{w}^*\|$.
    
    This means that, at any given round $t$, FedBaF's model weights are closer to $\mathbf{w}^*$.
\end{restatable}
 
Using the same restrictions on local and global learning rates placed by the FedAvg convergence analysis in \citep{karimireddy2020scaffold}, the aforementioned bounded gradient variance, bounded gradient dissimilarity, and  L-smoothness assumptions ensure that our method converges to $\mathbf{w}^*$ faster than FedAvg.
Similar convergence rate arguments for other FL methods with appropriately modified aggregation can be shown, as discussed in Sec.~\ref{sec: methodology}.

\subsection{Effectiveness of FedBaF with Diverse Client Data} \label{subsec:nonIID}
This section shows the impact of integrating a foundation model close to the optimal weights on the learning process and convergence behavior in non-IID settings.

In round $t$, client \(k\) uses multiple SGD steps to update the global model \(\mathbf{w}_t\)\ and obtains the local model \(\mathbf{w}_{t}^k\). 
Letting \(S_t\) represent the randomly selected set of active clients at time \(t\), we define \(\delta_t\) as the maximum deviation of the client models from \(\mathbf{w}^*\):
\begin{equation*}
    \delta_t := \max_{k \in S_t} \|\mathbf{w}_{t}^k - \mathbf{w}^*\|.
\end{equation*}
Aggregating the updated models from clients according to Alg.~\ref{algorithm: adapting pre-trained model} \textit{Lines 8-10} forms the global model \(\mathbf{w}'_{t}\). By the triangle inequality, we get
\begin{equation*}
    \|\mathbf{w}'_{t} - \mathbf{w}^*\| \leq \delta_t.
\end{equation*}

\textbf{Assumption: Foundation Model Proximity.} The foundation model's pre-trained weights, \(\mathbf{w}_{\text{pre}}\), are close to the optimal weights \(\mathbf{w}^*\), i.e., \(\|\mathbf{w}_{\text{pre}} - \mathbf{w}^*\| \leq \gamma\) for a small \(\gamma>0\). Furthermore, we assume that $\gamma \leq \delta_t$ for earlier rounds (small $t$ which makes $\tau_t\gg 0$), i.e., the foundation model is closer to the optimal model than clients' local weights.

These are reasonable assumptions in practice since selecting a foundation model with a large $\gamma$ would correspond to selecting an unsuitable foundation model that hampers the training process.

\begin{restatable}{proposition}{divergenceprop}
\label{prop:divergence}
Let $\mathbf{w}^*$ be a (bounded) local minimum of the global objective function in \eqref{eq:global_objective}. Consider an FL algorithm that converges to $\mathbf{w}^*$
and let $\mathbf{w}'_{t}$ be its global model. Consider FedBaF based on the same FL algorithm (with appropriately modified client updates and \textit{Lines 8-10} in Alg.~\ref{algorithm: adapting pre-trained model}) and let $\mathbf{w}_{t}$ be the FedBaF global model. 
FedBaF's global model error has an upper bound of $\|\mathbf{w}_{t}-\mathbf{w}^*\| \leq \frac{\delta_t + \alpha_t\tau_t \gamma}{1 + \alpha_t\tau_t} < \delta_t$.
\end{restatable}

Similar to \eqref{eq:squared_bound} in Sec.~\ref{subsec:convergence}, we bounded the distance between the FedBaF global model $\mathbf{w}_{t}$ and $\mathbf{w}^*$ in terms of $\delta_t$.
Prop.~\ref{prop:divergence} shows that the integration of the foundation model not only helps in stabilizing the learning process but also accelerates the convergence rate. The foundation model acts as a stabilizing factor and reduces the impact of this variance on the global model's convergence. This is particularly significant in the early stages of learning with non-IID data, when local models' weights are more prone to diverge from each other.
\section{Experimental Evaluations} \label{sec: experiments}
\begin{figure*}[ht]
    \centering
    \includegraphics[width=\linewidth]{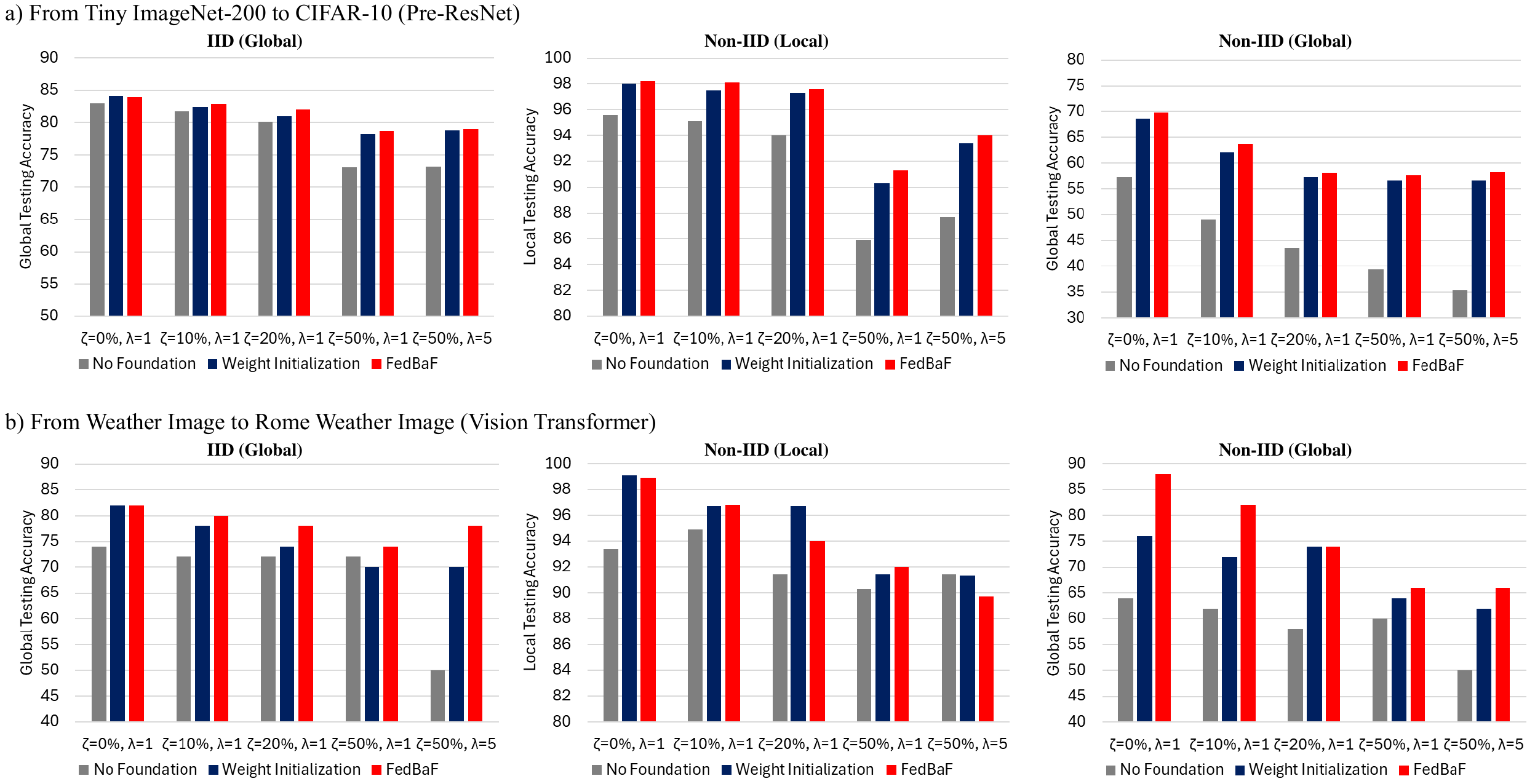}
    \vspace{-0.5 cm}
    \caption{FedBaF maintains higher test accuracy when used with \textbf{FedAvg} with different proportions of malicious clients, $\zeta$ (0\%, 10\%, 20\%, 50\%), and attack intensity, $\lambda$ (1, 5), executing misclassification attacks, under IID and non-IID settings. Three different foundation models, trained with different datasets, are used for three tasks. Red, blue, and gray bars respectively represent FedBaF, weight initialization, and no foundation model cases.}
    \label{fig:attack curve case 3 fedavg}
\end{figure*}
\begin{figure*}[ht]
    \centering
    \includegraphics[width=\linewidth]{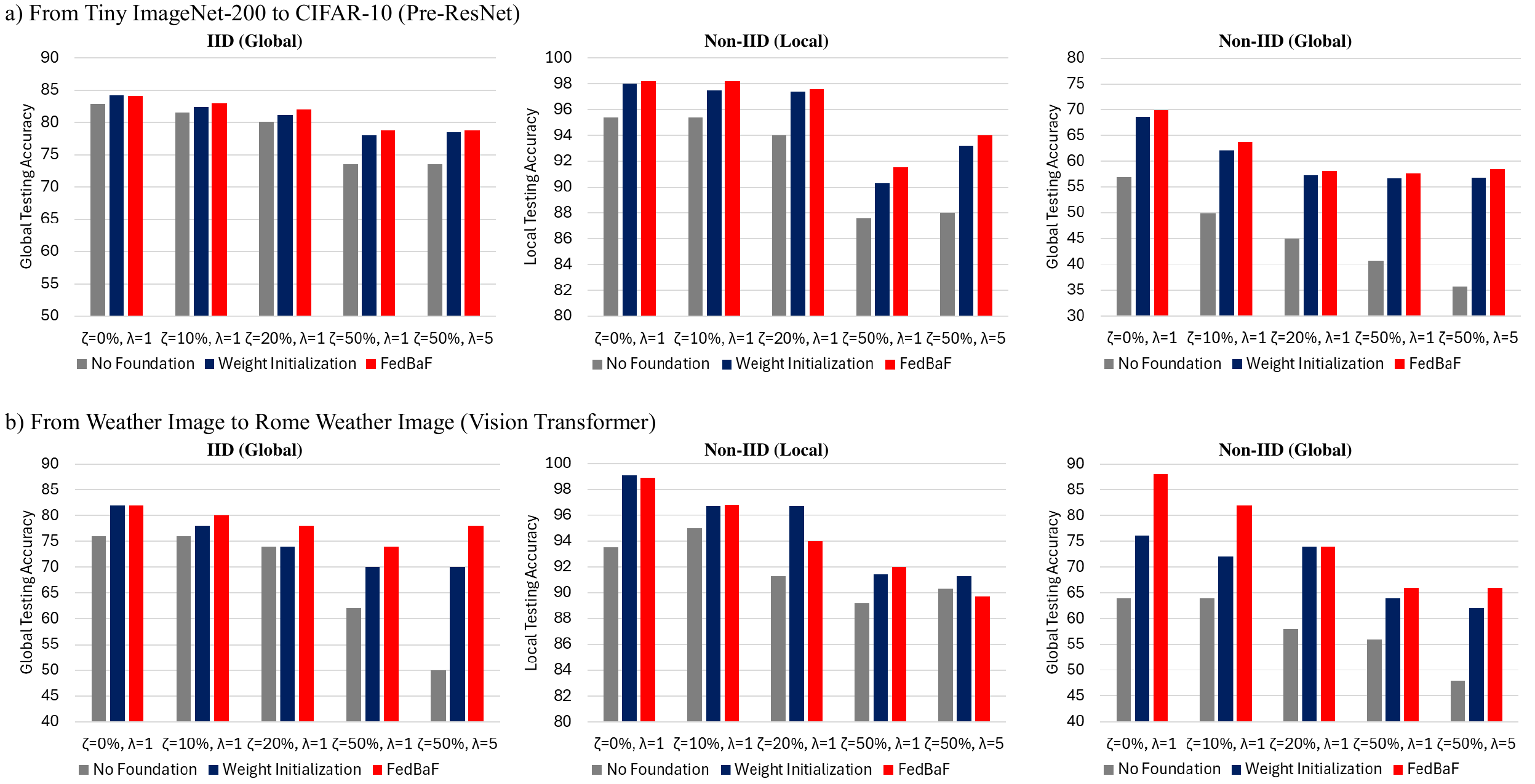}
    \vspace{-0.5 cm}
    \caption{FedBaF maintains higher test accuracy than both baselines when used with \textbf{FedProx} with different proportions of malicious clients, attack intensity, and IID as well as non-IID settings. All other settings are identical to those in Figure~\ref{fig:attack curve case 3 fedavg}.}
    \label{fig:attack curve case 3 fedprox}
\end{figure*}
\begin{figure}[ht]
    \centering
    \includegraphics[width=5.5 cm]{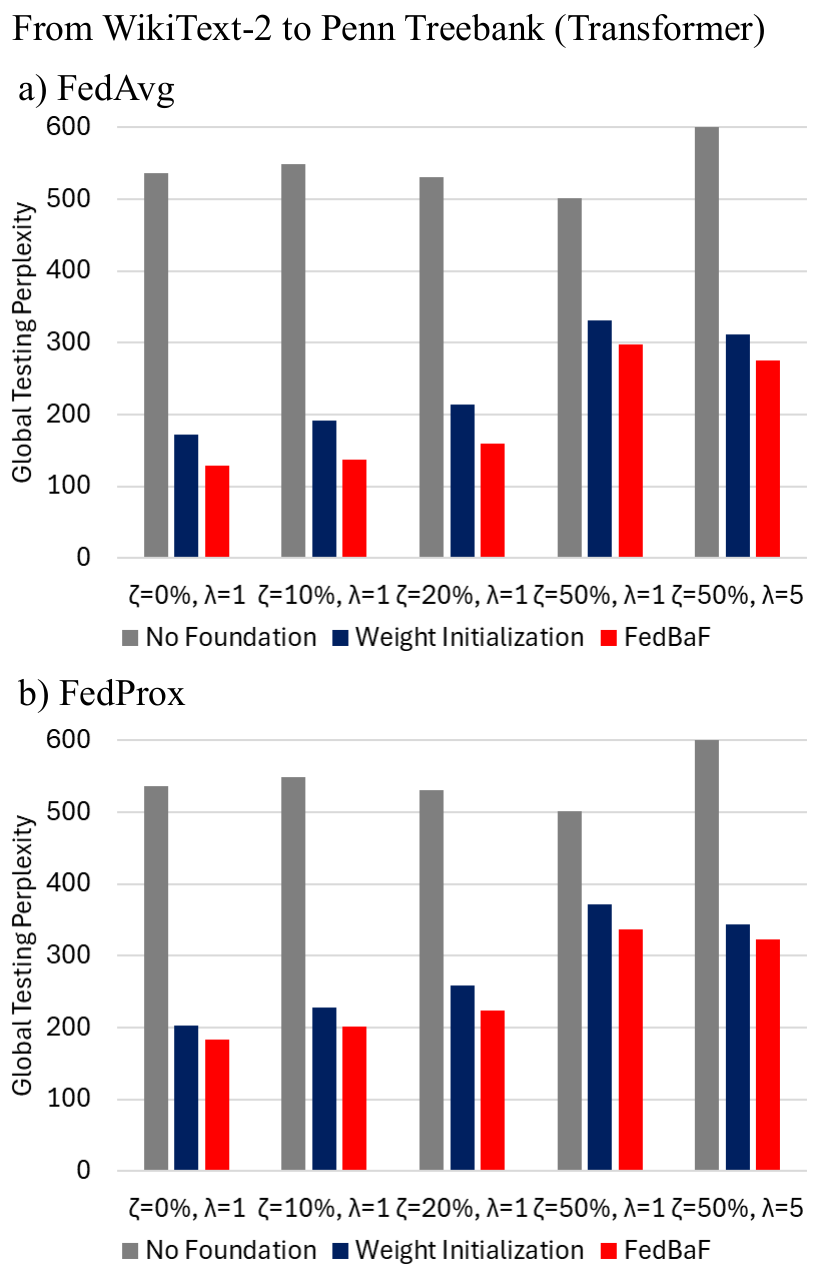}
    \vspace{-0.5 cm}
    \caption{FedBaF maintains lower test perplexity when used with \textbf{FedAvg} and \textbf{FedProx} with different proportions of malicious clients and attack intensities.
    \textbf{Note: lower perplexity is better}}
    \label{fig:attack curve case 3 language}
\end{figure}

In this section, we conduct a detailed evaluation of FedBaF's performance on both local and global test datasets, comparing its performance to the \textbf{no foundation model} and \textbf{weight initialization} baseline algorithms for training FL models. %In this context, 

$\bullet$ \textbf{No foundation model}: The global FL model is trained from scratch without weight initialization or FedBaF (i.e., no use of foundation models).

$\bullet$ \textbf{Weight initialization}: The global model's initial weights are set to equal the foundation model's weights, and the FL training then proceeds as usual. 

We aim to \textbf{1)} illustrate that FedBaF offers security advantages over weight initialization while attaining equivalent performance. 
Furthermore, by verifying how $\tau_t$ in line 9 of Algorithm~\ref{algorithm: adapting pre-trained model} converges to 0, we also \textbf{2)} establish FedBaF's ability to effectively adapt the influence of the foundation model.

More detailed testing results are provided in Appendix~\ref{appendix:additional_experimental_evaluation}, including the \textbf{1)} computational efficiency of FedBaF and \textbf{2)} additional evaluations using foundation models of varying quality trained on different amounts of data and real-world foundation model weights that are publicly available online.

\subsection{Experimental Setup}
Our experiments with popular image classification tasks encompass experiments using the CIFAR-10 and Rome Weather Image~\citep{vaz_rome_weather_classification} datasets. We use Pre-ResNet and Vision Transformer~\citep{dosovitskiy2020image} architectures as Vision Transformers are popular foundation model architectures known for achieving remarkable performance~\citep{zhou2023comprehensive} for ImageNet challenges~\citep{dosovitskiy2020image}. We train the foundation models on the Tiny ImageNet-200 and Weather Image~\citep{DVN/M8JQCR_2021} datasets. 
Our evaluations consider both IID and non-IID settings (see Appendix~\ref{appendix:experimental_setup}).
To demonstrate FedBaF's generalizability to other tasks, we also evaluate it on a next-word prediction task using a Transformer language model pre-trained on the WikiText-2 dataset and tested on the Penn Treebank dataset. 

\textbf{Attack setup.} To assess security robustness of FedBaF, we randomly shuffle local data labels for a subset of clients, treating them as backdoor attackers aiming to induce misclassification. We also increase the attack intensity by varying the number of local epochs for malicious clients. For image classification tasks (Pre-ResNet and Vision Transformer), we increase the local epochs by a factor $\lambda > 1$, which introduces more bias from the initial global model and strengthens the attack's impact. For the language task (Transformer), we decrease the local epochs by a factor $1/\lambda < 1$ to prevent convergence to a small loss, ensuring the calculated perplexity remains high regardless of misclassification, thereby intensifying the attack. 
We vary the proportion of attacking clients, $\zeta$, and evaluate algorithm resiliency when 0\%, 10\%, 20\%, and 50\% of the client base are attackers.

\textbf{Evaluation metrics.} To evaluate testing performance, we calculate the \textit{global testing accuracy} using a global test dataset after the aggregation phase of an FL round. In non-IID settings, we also use local test datasets that are extracted from the global test dataset and reflect the class distribution of the local clients. After local training and prior to aggregation, we test the local models to determine an \textit{average local testing accuracy}. 
For the Transformer model, we use \textit{global perplexity} to assess the performance of the global language model. Perplexity is \textbf{inversely} related to how well a probability model predicts a sample.

Details of other deep neural network architectures employed in our experiments and additional training specifics are provided in Tables~\ref{tab:test condtion} and~\ref{tab:network architecture} in Appendix~\ref{appendix:experimental_setup}.

\subsection{Experimental Results}
Figures~\ref{fig:attack curve case 3 fedavg}, \ref{fig:attack curve case 3 fedprox}, and \ref{fig:attack curve case 3 language} display the extensive test accuracy/perplexity evaluation results for FedBaF and our two baselines (\textit{no foundation model} and \textit{weight initialization}).
We evaluate the three methods using both FedAvg and FedProx~\citep{li2020federated} as the base FL training algorithms. We incorporate FedBaF into FedProx by modifying \textit{Line 8} and the ClientUpdate routine in Algorithm~\ref{algorithm: adapting pre-trained model}, including both IID and non-IID settings, with one non-adversarial scenario and four adversarial scenarios. We use FedProx's aggregation step: \(\mathbf{w}'_{t+1} \gets \sum_{k \in S_t} \frac{n_k}{\sum_{k \in S_t} n_k} \mathbf{w}_{t+1}^k - \mu (\mathbf{w}_{t+1}^k - \mathbf{w}_t)\). Here, \(\mu\) represents the regularization term that controls the trade-off between the local and global objectives. FedBaF then incorporates the foundation model weights as in \textit{Lines 9-11} of Algorithm~\ref{algorithm: adapting pre-trained model}.

\subsubsection{Testing FedBaF Performance}
\textit{In non-adversarial scenarios, FedBaF showcased superior testing performance compared to the no foundation model and weight initialization methods across both IID and non-IID configurations.}

Figures~\ref{fig:attack curve case 3 fedavg} and~\ref{fig:attack curve case 3 fedprox} respectively show test accuracies of all three methods using FedAvg and FedProx for both Pre-ResNet and the Vision Transformer.
In comparison to Pre-ResNet trained with no foundation model, FedBaF improved global model accuracy by 1.3\% for FedAvg and 1.6\% for FedProx in IID scenarios, and by 21.8\% for FedAvg and 22.6\% for FedProx in non-IID scenarios.
For the Vision Transformer, FedBaF improved global performance relative to no foundation model by 10.8\% for FedAvg and 0.0\% for FedProx in IID settings and by 37.5\% for FedAvg and 15.8\% for FedProx in non-IID settings. We observe that \textbf{both FedAvg and FedProx benefit from FedBaF's inclusion of the foundation model}, with particular benefits in more challenging scenarios with non-IID client data. Incorporating the foundation model mitigates slower FL convergence caused by non-IID data.
 
The weight initialization method, which also incorporates a foundation model but does not keep it private, exhibits similar performance gains as FedBaF compared to training without a foundation model. For example, on Pre-ResNet weight initialization exhibited global performance gains of 19.7\% for FedAvg and 20.4\% for FedProx in non-IID scenarios, while on Vision Transformers it achieves gains of 18.8\% for both FedAvg and FedProx in non-IID scenarios.

In the next-word prediction task using a Transformer, FedBaF significantly outperformed training with no foundation model, reducing perplexity by 76.0\% with FedAvg, whereas weight initialization yielded a 67.8\% decrease relative to training without a foundation model with FedAvg. 

Collectively, these findings indicate \textbf{negligible differences between the test accuracies attained by FedBaF and those achieved with weight initialization}. Simultaneously,\textbf{ FedBaF achieves privacy advantages} as, unlike weight initialization, it does not reveal the foundation model weights to FL clients. Moreover, FedBaF showed better testing performance than the weight initialization or no foundation model methods in 10 out of the 14 experiment settings. 

In Appendix~\ref{appendix:adam_results}, we provide additional examples of FedBaF achieving privacy and test performance improvements. They include comparisons with FedAdam~\citep{reddiadaptive}, a state-of-the-art FL method leveraging pre-trained models.

\subsubsection{FedBaF's Robustness to Attacks}
\textit{FedBaF remains effective in maintaining robustness when faced with misclassification attacks and shows a more modest performance degradation in comparison to both baseline methods.}

We evaluate the robustness of FedBaF and our two baselines in the presence of adversarial clients. As the proportion of attacking clients ($\zeta$ in Figures~\ref{fig:attack curve case 3 fedavg},~\ref{fig:attack curve case 3 fedprox}, and~\ref{fig:attack curve case 3 language}) increases, test accuracy declines in all cases. This matches our intuition since neither FedBaF nor the two baselines are designed to perfectly defend against these attacks, which become more effective as more clients act as attackers. 

The test accuracy for the no foundation model baseline method drops significantly when 50\% of the clients are attackers and the attack intensity $\lambda = 5$. For Pre-ResNet under such attacks, the global performance drop, when compared to the case with no attackers, is 11.8\% for FedAvg and 11.2\% for FedProx in IID scenarios and 38.2\% for FedAvg and 37.4\% for FedProx in non-IID scenarios.
With the Vision Transformer, these accuracy drops are 32.4\% for FedAvg and 34.2\% for FedProx in IID and 21.9\% for FedAvg and 25.0\% for FedProx in non-IID scenarios. Thus, \textbf{training with no foundation model results in vulnerability to attacks in both IID and non-IID settings}. This matches our expectations since it has no built-in defenses.

In contrast, \textbf{FedBaF experiences a much more modest performance decline under attack}, demonstrating its robustness. For Pre-ResNet, FedBaF's global performance decreases by 6.0\% for FedAvg and 6.4\% for FedProx in IID settings and by 16.5\% for FedAvg and 16.3\% for FedProx in non-IID settings, where the decrease is again measured for the most intense attack ($\zeta = 50\%, \lambda = 5$) relative to no attack.
These accuracy drops are \textit{less than half} of those experienced by the no foundation model method.
With the Vision Transformer, FedBaF's global performance decreases by only 4.9\% for FedAvg and FedProx in IID settings and by 25\% for both FedAvg and FedProx in non-IID settings. 

Finally, we comment that the results also show that the weight initialization method, which na\"ively incorporates the foundation model and reveals the foundation model weights to clients, does confer some robustness to attacks. Weight initialization shows a performance decrease of 6.4\% for FedAvg and 6.9\% for FedProx in IID settings and 17.5\% for FedAvg and 17.2\% for FedProx in non-IID settings globally for Pre-ResNet. With the Vision Transformer, the decreases are 14.6\% for both FedAvg and FedProx in IID settings and 18.4\% for both FedAvg and FedProx in non-IID settings globally.

These findings demonstrate that, similar to weight initialization, \textbf{FedBaF offers considerable attack robustness} compared to training with no foundation model. 
In 8 out of 12 cases, FedBaF suffers the minimal loss in performance when compared with the other two baselines, indicating its superior effectiveness in maintaining robustness under adversarial conditions.

\section{Conclusion} \label{sec: conclusion}
This paper introduced Federated Learning Aggregation Biased by a Foundation Model (FedBaF). FedBaF enhances adaptability and security in dynamic FL scenarios without sharing the foundation model with clients. This is crucial in environments with ever-changing data and non-IID scenarios, where foundation models are used across several domains as seed models. Our findings show that FedBaF increases resilience against adversarial attacks while matching or outperforming traditional weight initialization performance in both IID and non-IID settings.

\newpage
\bibliographystyle{alpha}
\bibliography{bibliography}

\appendix
\onecolumn
\clearpage
\newpage
\section*{Appendix Overview}
This appendix provides additional details and results to complement the main text, offering further insight into the experimental setup, theoretical analysis, and security considerations for FedBaF. The appendix is organized as follows:

\subsection*{Experimental Setup in Section~\ref{appendix:experimental_setup}}
This section details the experimental configurations, including data distributions, network architectures for vision and language tasks, and hyper-parameters under IID and non-IID settings.

\subsection*{Formulas for Evaluating Computational Complexity in Section~\ref{appendix:comp_complexity}}
We present the mathematical formulas used to compute the computational complexity of the FedBaF algorithm, specifically focusing on multiply-accumulate (MAC) operations across clients and training rounds.

\subsection*{Additional Experimental Evaluations in Section~\ref{appendix:additional_experimental_evaluation}}
This section includes supplementary experimental results, analyzing the effect of varying foundation model quality on FedBaF’s performance and comparisons using the official pre-trained foundation model. Computational complexity is also compared to scenarios without foundation models and weight initialization. Additionally, we include further experiments employing FedAdam~\citep{reddiadaptive}.

\subsection*{Training Curves in Section~\ref{appendix:training_curves}}
We provide training curves that display the progression of model accuracy over training epochs for the Pre-ResNet and Vision Transformer models, illustrating comparisons between FedBaF, weight initialization, and without foundation model cases under IID and non-IID conditions.

\subsection*{Convergence Analysis in Section~\ref{appendix:theory}}
This section provides a theoretical analysis of FedBaF's convergence properties, detailing how the algorithm performs under different client data distributions and demonstrating the theoretical guarantees for its performance.

\subsection*{Proofs of Propositions in Section~\ref{appendix:proposition}}
In this section, we include the formal proofs of Propositions~\ref{prop:improvement}
 and \ref{prop:divergence}, outlined in the theoretical analysis in Section~\ref{sec: analysis}, which support the claims made regarding FedBaF’s performance and convergence behavior.

\subsection*{Security Analysis in the Presence of Adversarial Attacks in Section~\ref{appendix:security}}
This section provides an in-depth security analysis of FedBaF, focusing on its robustness against adversarial attacks such as misclassification and backdoor attacks. We compare FedBaF’s resilience to malicious clients with traditional methods, showing how FedBaF mitigates the negative impact of attacks and preserves global model integrity.
\clearpage
\newpage
\section{Experimental Setup} \label{appendix:experimental_setup}
\begin{table*}[ht]
    \centering
    \caption{The specific conditions under which our experiments were conducted, including data distribution and model training settings.}
    \includegraphics[width=\linewidth]{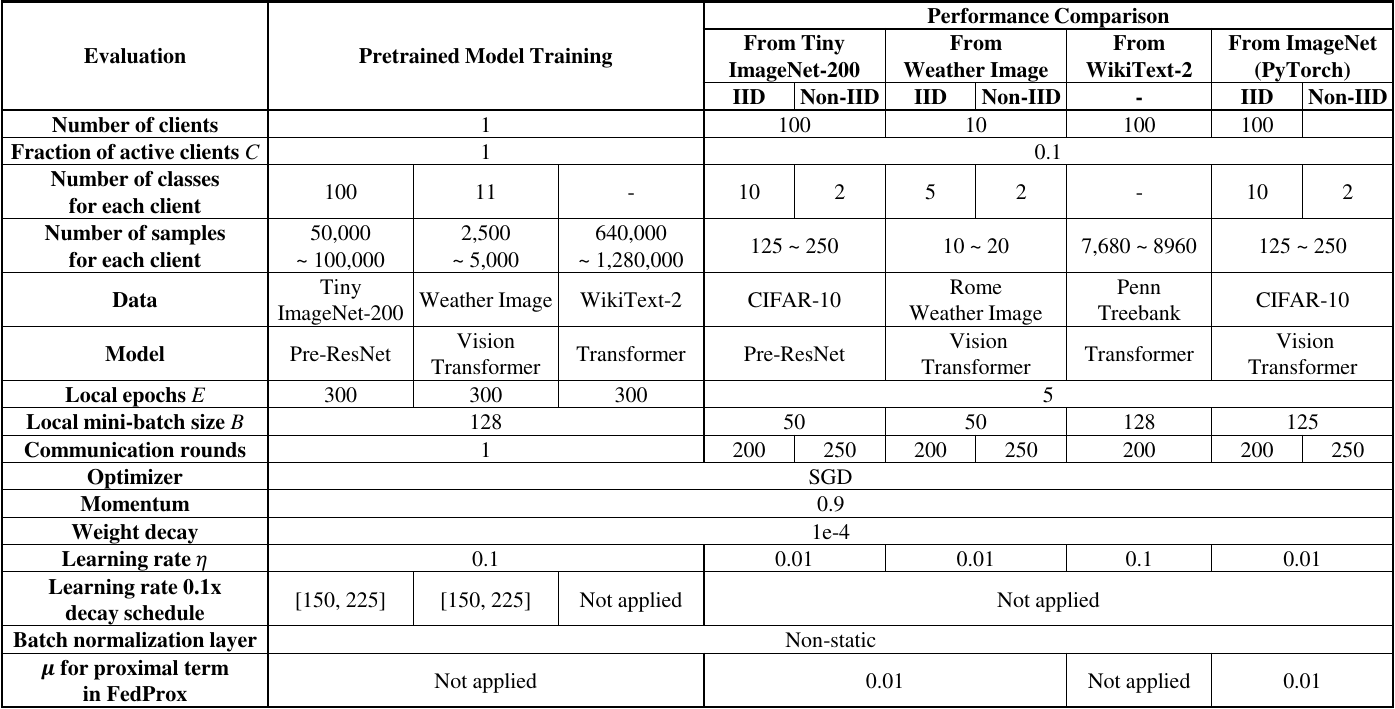}
    \label{tab:test condtion}
\end{table*}
In this section, we provide details about our experimental setup and network architectures. For vision tasks using Pre-ResNets and Vision Transformers, we conduct evaluations in both IID and non-IID environments. In \textbf{IID settings}, each client's data distribution is uniform across all classes, with an equal number of samples from each class. In \textbf{non-IID settings}, clients receive samples from only 20\% of the dataset's classes for CIFAR-10 and 50\% of the dataset's classes for Rome Weather Image, but maintain an equal number of samples for each class they have.
\textit{During local training in these settings, clients zero out logits for classes not present in their data.} For language tasks using Transformers, each client has specific numbers of tokenized words grouped sequentially. Details of the experimental settings can be found in Table~\ref{tab:test condtion}.

For the network architecture configuration, details for Pre-ResNets and Transformers can be found in Table~\ref{tab:network architecture}. For Vision Transformer, we used the standard ViT$\_$B$\_$16 model with no modifications except changing the last output layers according to the new data. This model was obtained from the PyTorch library. Additionally, we also tested FedBaF with official pre-trained foundation model weights that are available online (not developed by us). We specifically used the ImageNet pre-trained weights for a standard Vision Transformer from PyTorch's official model repository: (ViT$\_$B$\_$16$\_$Weights.IMAGENET1K$\_$SWAG$\_$E2E$\_$V1).

\begin{table*}[ht]
    \centering
    \caption{This table presents the detailed structures of the neural networks (Pre-ResNet and Transformer) utilized in our Federated Learning experiments and making foundation models.}
    \includegraphics[width=\linewidth]{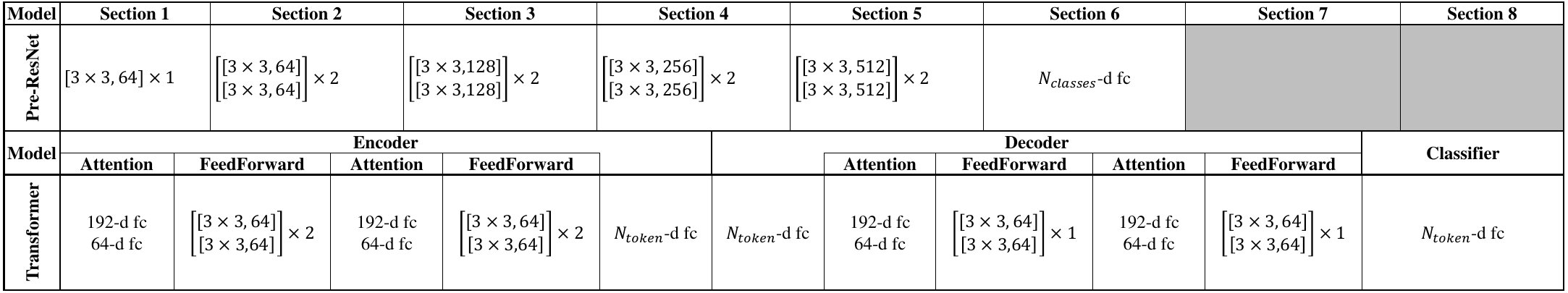}
    \label{tab:network architecture}
\end{table*}

\newpage
\clearpage
\section{Formulas for Evaluating Computational Complexity}\label{appendix:comp_complexity}
In this section, we provide metrics for evaluating the computational complexity, as discussed in Section~\ref{subsec: comp_complexity}.

To evaluate \textit{computational complexity}, we track the number of multiply-accumulate (MAC) operations, denoted as $\text{MACS}$. $\text{MACS}_k$ represents the number of MAC operations required to process all the local data samples held by client $k$. The average MAC per client, denoted as $\overline{\text{MACS}}$, is calculated by averaging the $\text{MACS}$ values for all clients participating in a single round of FL:

\begin{equation*}
    \overline{\text{MACS}} = \frac{1}{m} \sum_{k=1}^{m} \text{MACS}_k
\end{equation*}

Here, $m$ is the number of participating clients in the given round. For each local training epoch, the computational complexity, referred to as $\text{MACE}$ (Multiply-Accumulate Complexity per Epoch), is calculated as:

\begin{equation*}
    \text{MACE} = m \times \tilde{n} \times \overline{\text{MACS}}.
\end{equation*}
Here, $\tilde{n}$ is the median number of data samples per client. This metric reflects the computational load incurred by $m$ clients during local training in each epoch.

To compute the total computational load for the entire FL system, denoted as $\text{TMAC}$ (Total Multiply-Accumulate Complexity), we multiply the number of local epochs $E$, the number of aggregation rounds $T$, and the previously calculated $\text{MACE}$:

\begin{equation*}
    \text{TMAC} = T \times E \times \text{MACE}.
\end{equation*}

This provides the total number of MAC operations required for the entire training process across all clients and rounds, accounting for both the number of local epochs and the aggregation rounds in the FL system.

\clearpage
\newpage
\section{Additional Experimental Evaluations} 
\label{appendix:additional_experimental_evaluation}
In this section, we present additional experimental results that provide further insight into the performance of FedBaF across various tasks and scenarios. These evaluations focus on the impact of different qualities of foundation models, the application of the real pre-trained foundation model, and computational complexities. We compare the use of foundation models in both IID and non-IID settings with weight initialization and cases where no foundation model is used.
\newpage
\subsection{Differentiating the Quality of Pre-Trained Foundation Models}
We conducted experiments to evaluate the generalized performance of FedBaF by varying the quality of foundation models. Tables~\ref{tab:test result resnet}, \ref{tab:test result vit}, and \ref{tab:test result transformer} present the results for image classification tasks using Pre-ResNet and Vision Transformer models, as well as a next-word prediction task using a Transformer model.

\begin{table}[H]
    \centering
    \caption{Image classification test accuracy results for Pre-ResNet using the no foundation, weight initialization, and FedBaF methods (best of 3 trials).}
    \includegraphics[width=\linewidth]{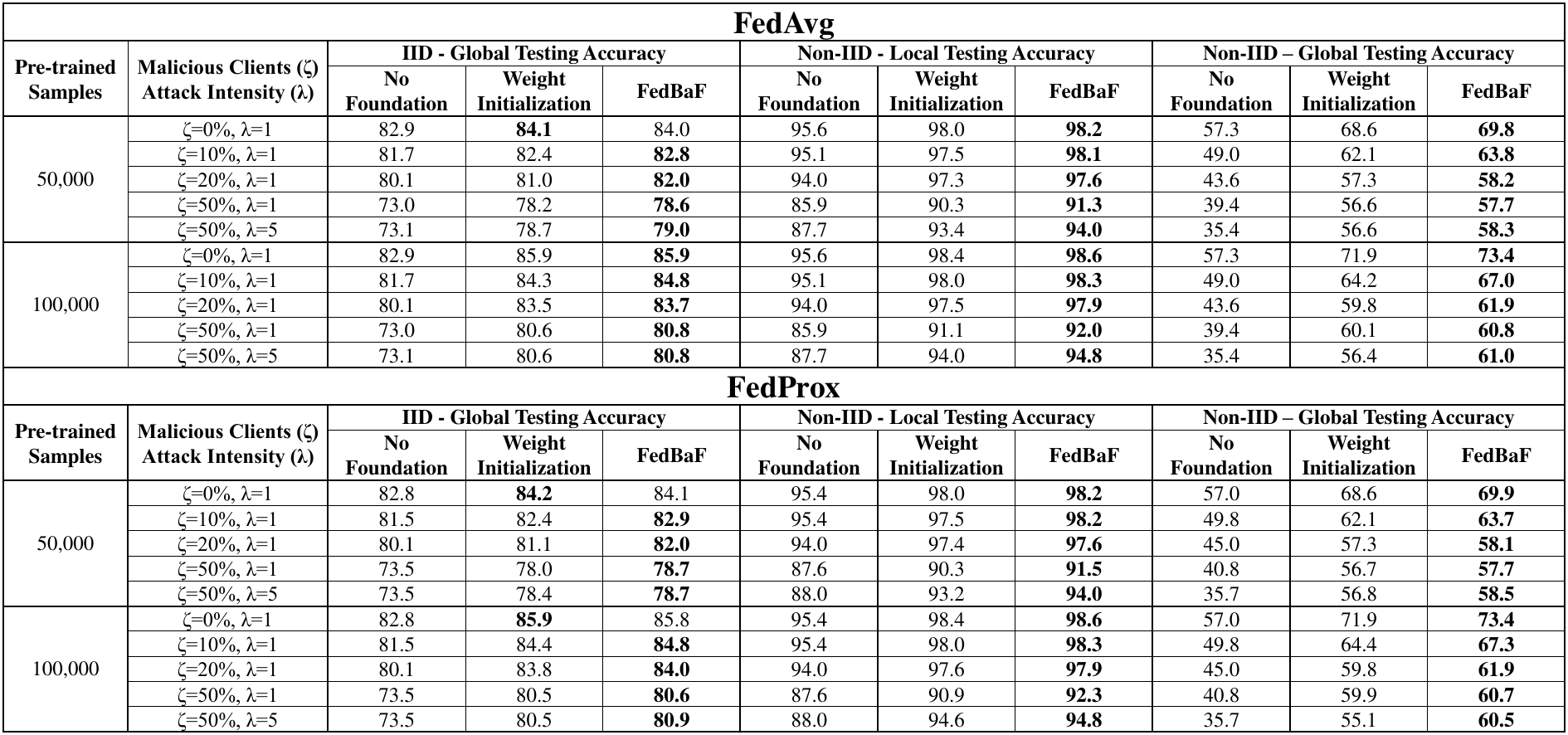}
    \label{tab:test result resnet}
\end{table}
\begin{table}[H]
    \centering
    \caption{Image classification test accuracy results for Vision Transformer using no foundation, weight initialization, and FedBaF methods (best of 3 trials).}
    \includegraphics[width=\linewidth]{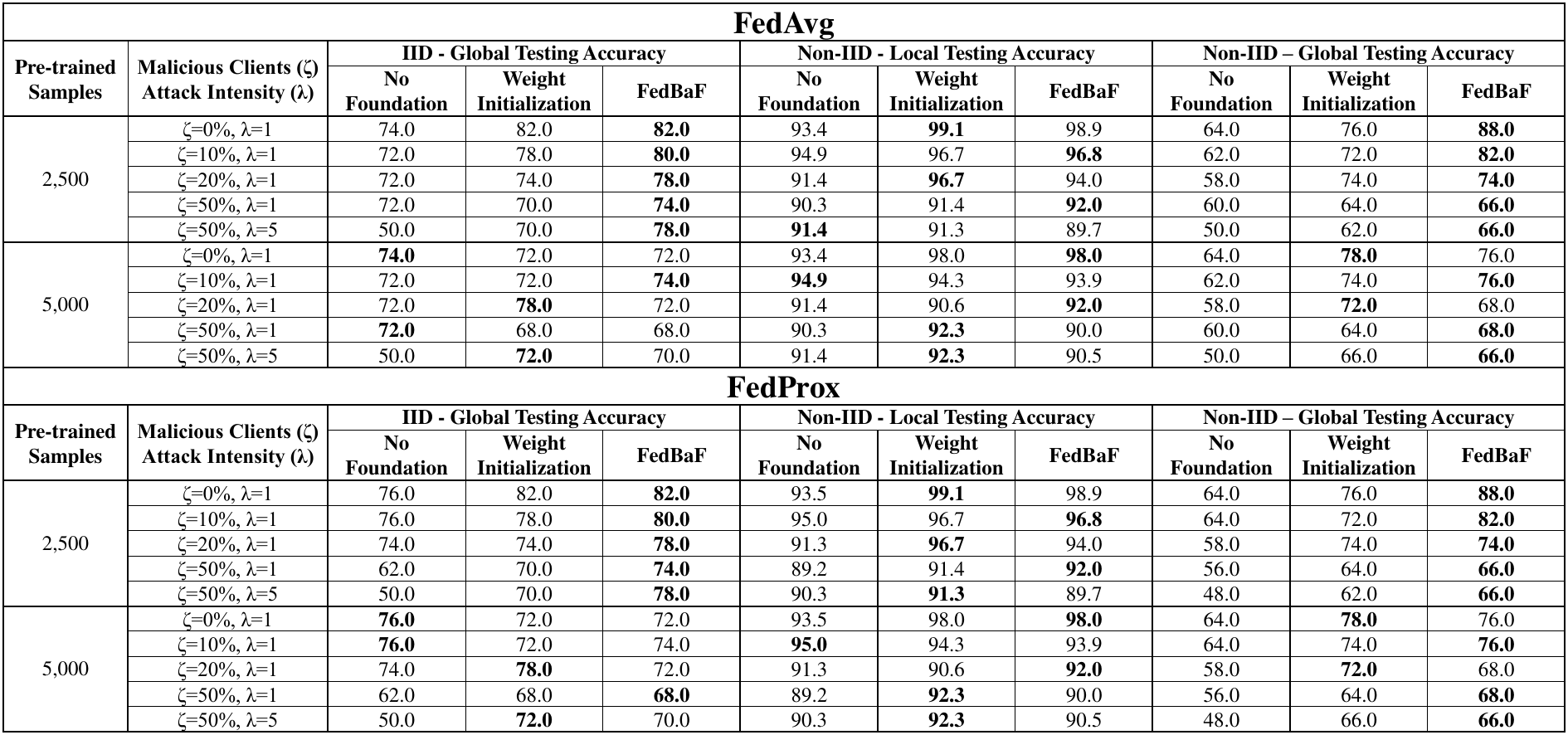}
    \label{tab:test result vit}
\end{table}

To assess the impact of foundation model quality, we varied the number of pre-trained samples used for each model and assessed each method's performance under a varying number of malicious clients and attack intensity. Interestingly, larger sample sizes do not always lead to better results, as seen in Table~\ref{tab:test result resnet}, where excessive pre-training can negatively impact performance. This trend is further evidenced in Tables~\ref{tab:test result vit}, \ref{tab:test result transformer}. The reason behind this is likely due to overfitting or reduced adaptability to new tasks. Despite these variations, FedBaF consistently outperforms models without foundation models and delivers similar testing performance to weight initialization.

Training curves for selected cases can be found in Section~\ref{appendix:training_curves}.

\begin{table}[H]
    \centering
    \caption{Next-word prediction perplexity results for Transformer models using no foundation, weight initialization, and FedBaF methods (best of 3 trials). \textbf{Lower perplexity is better.}}
    \includegraphics[width=\linewidth]{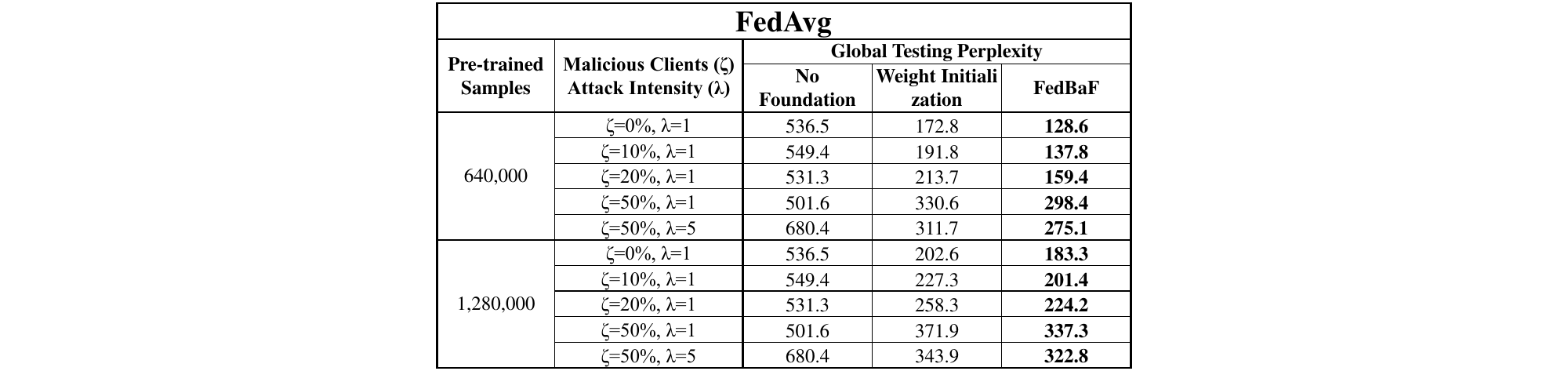}
    \label{tab:test result transformer}
\end{table}

\newpage
\subsection{Using the Official Pre-Trained Foundation Model}
\begin{table}[ht]
    \centering
    \caption{Image classification test accuracy results for Vision Transformer using official pre-trained foundation model weights from PyTorch. Comparisons are made between no foundation model, weight initialization, and FedBaF methods (best of 3 trials).}
    \includegraphics[width=\linewidth]{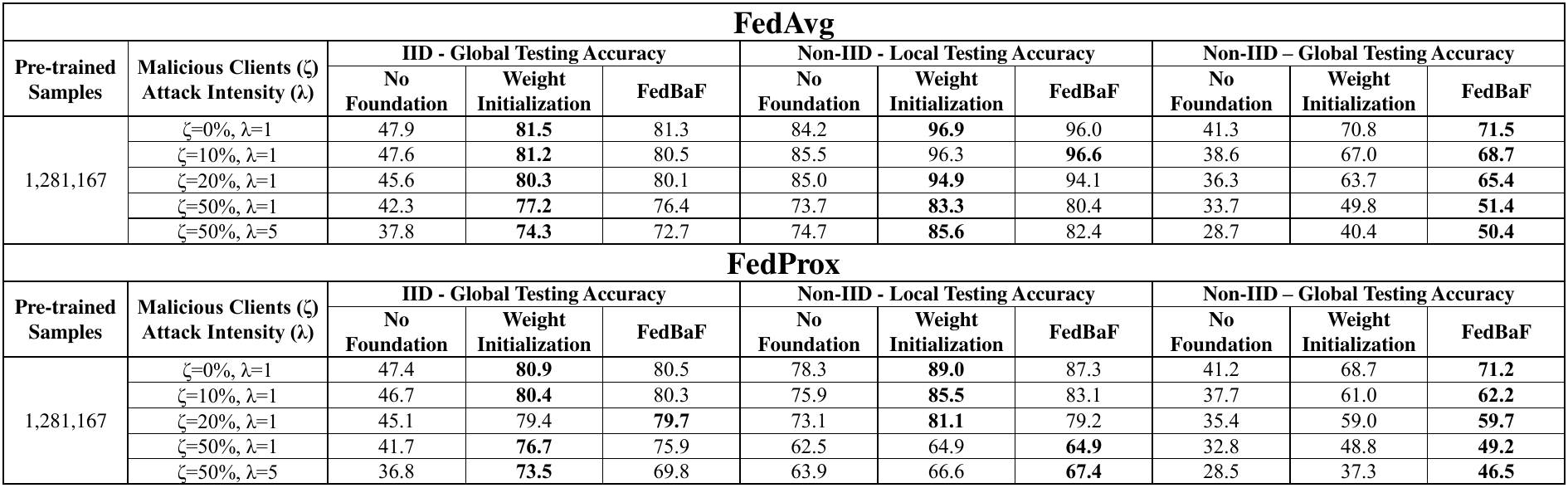}
    \label{tab:test result real}
\end{table}

We also evaluated the performance of FedBaF using the official pre-trained foundation model that was not developed by us. Specifically, we used ImageNet pre-trained weights for the Vision Transformer model, obtained from PyTorch’s official model repository. As shown in Table~\ref{tab:test result real}, FedBaF outperforms scenarios without foundation models and delivers similar performance to weight initialization using pre-trained weights. These results demonstrate that FedBaF can effectively integrate widely adopted pre-trained models.

\newpage
\subsection{Generalized Applicability of FedBaF with FedAdam} \label{appendix:adam_results}
To further assess the generalized applicability of FedBaF, we conducted additional experiments using the FedAdam~\cite{reddiadaptive} algorithm. Specifically, we evaluated its performance on image classification tasks with CIFAR-10 and Rome Weather Image datasets using Pre-ResNet and Vision Transformer, as well as on a language modeling task with a Transformer model pre-trained on the WikiText-2 dataset. We followed the same experimental setup, attack scenarios, and hyperparameters described in Section~\ref{sec: experiments} and Appendix~\ref{appendix:experimental_setup}.

For FedAdam, we set the global aggregation update learning rate to 0.01 for CIFAR-10 and Rome Weather Image datasets and 0.1 for the WikiText-2 dataset. Additionally, we configured the momentum parameters as $\beta_1 = 0.9$ and $\beta_2 = 0.99$ for all experiments.

\begin{table}[H]
    \centering
    \caption{Image classification test accuracy results for Pre-ResNet using the no foundation, weight initialization, and FedBaF methods employing \textbf{FedAdam} (best of 3 trials).}
    \includegraphics[width=\linewidth]{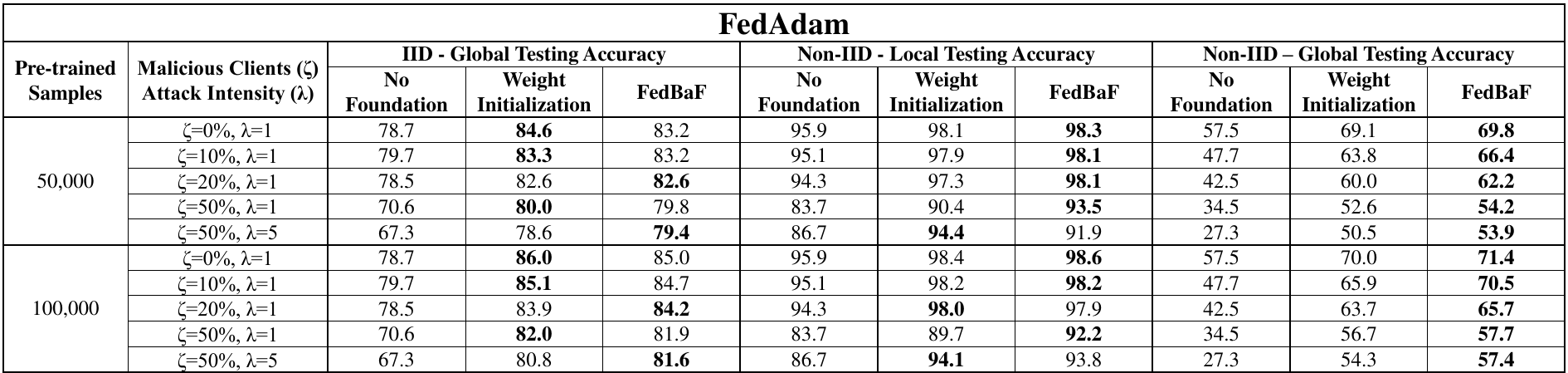}
    \label{tab:test result cifar10 adam}
\end{table}
\begin{table}[H]
    \centering
    \caption{Image classification test accuracy results for Vision Transformer using no foundation, weight initialization, and FedBaF methods employing \textbf{FedAdam} (best of 3 trials).}
    \includegraphics[width=\linewidth]{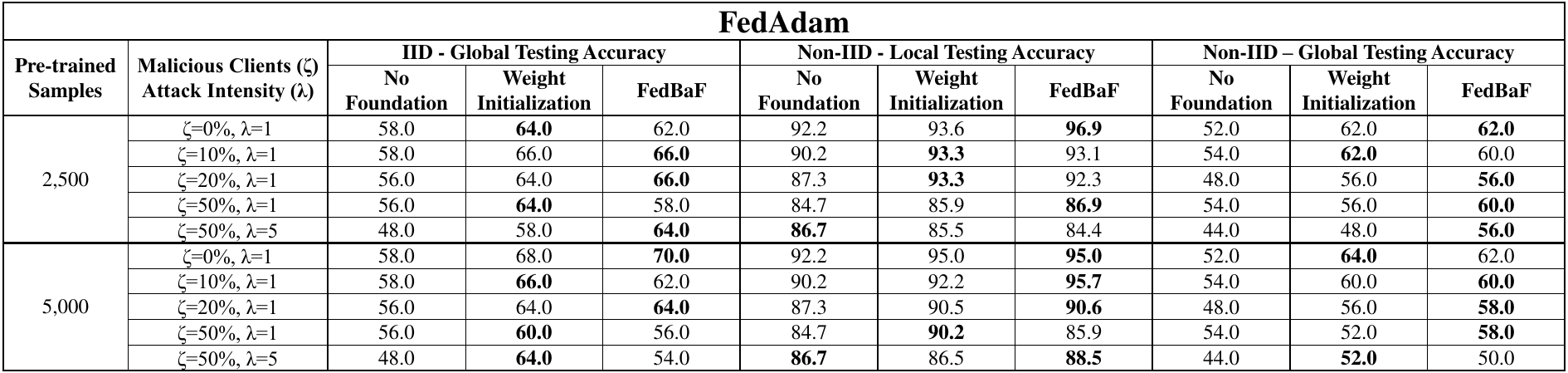}
    \label{tab:test result weather adam}
\end{table}
\begin{table}[H]
    \centering
    \caption{Next-word prediction perplexity results for Transformer models using no foundation, weight initialization, and FedBaF methods employing \textbf{FedAdam} (best of 3 trials). \textbf{Lower perplexity is better.}}
    \includegraphics[width=\linewidth]{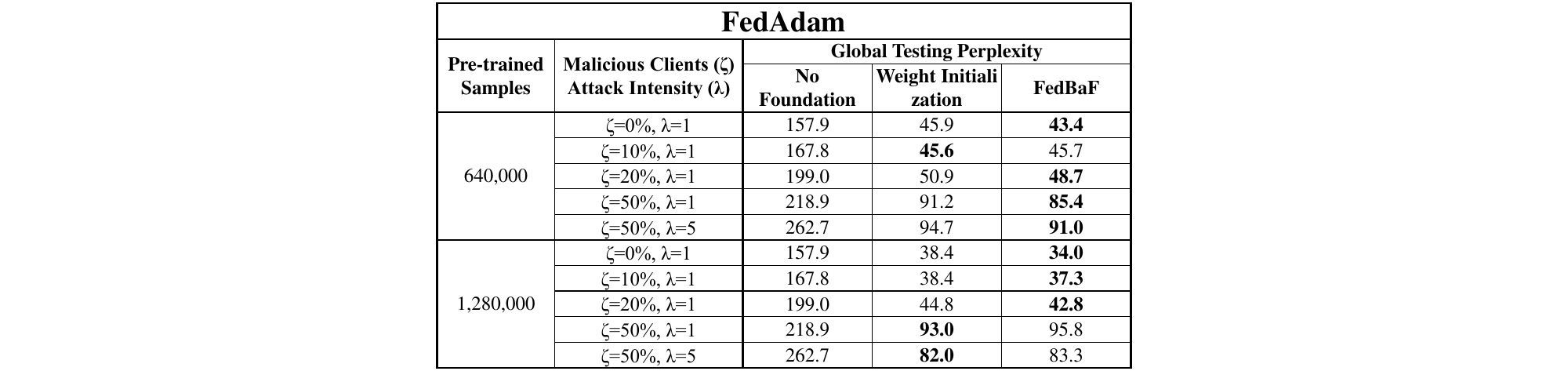}
    \label{tab:test result transformer adam}
\end{table}

Tables~\ref{tab:test result cifar10 adam}, \ref{tab:test result weather adam}, and \ref{tab:test result transformer adam} present the experimental results obtained using FedAdam across different datasets, model architectures, and attack scenarios. The results demonstrate that FedBaF remains effective across various federated optimization frameworks, reinforcing its adaptability in federated learning.

Among the 70 tested cases, FedBaF achieved similar or superior performance in 68 cases when compared to weight initialization approaches, outperforming cases without foundation models. This trend aligns with the findings in Section~\ref{sec: experiments}, further highlighting FedBaF's robustness across different optimization strategies and experimental settings.

These findings confirm the reliability of FedBaF in federated learning, ensuring its applicability to both vision and language tasks under diverse conditions.

\clearpage
\newpage
\subsection{Computational Complexity} \label{subsec: comp_complexity}
\begin{table}[ht]
    \centering
    \caption{Pre-ResNet and Vision Transformer computational complexities using no foundation model, weight initialization, and FedBaF methods. \textbf{Note: T represents trillion.}}
    \includegraphics[width=\linewidth]{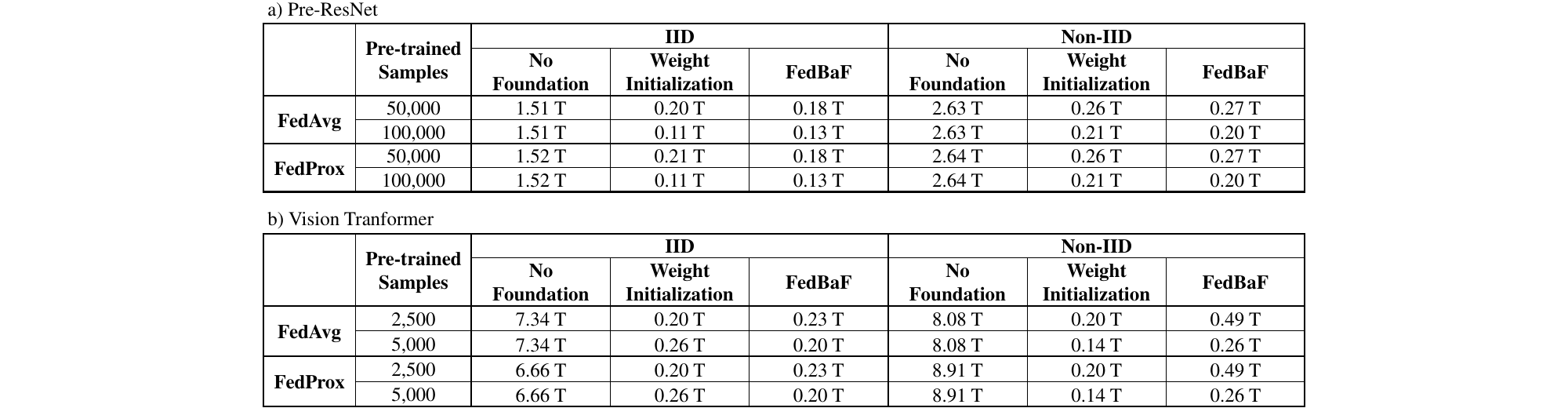}
    \label{tab:computational result}
\end{table}

\textit{FedBaF demonstrates remarkable efficiency in terms of computational complexity, requiring significantly fewer computations than scenarios without foundation models and performing similarly to weight initialization methods.}

To demonstrate that FedBaF’s computational demands are minimal, even when integrating the foundation model in every training round, we assess its computational complexity. Table~\ref{tab:computational result} presents the computational complexities, measured in TMAC (Total Multiply-Accumulate Operations) from Section~\ref{appendix:comp_complexity}, for six non-adversarial scenarios in both IID and non-IID settings. To calculate these complexities, we consider the number of training rounds required to achieve specific global testing accuracies: 75\% for IID and 50\% for non-IID scenarios in Pre-ResNet cases. For Vision Transformer cases, we set the thresholds to 60\% IID accuracy and 60\% non-IID accuracy. 

Compared to the no foundation model cases, \textbf{FedBaF requires significantly fewer computations} across both IID and non-IID scenarios. Specifically, for Pre-ResNet IID and non-IID cases, computations are reduced by 88.1-91.4\% and 89.7-92.4\% for FedAvg, and by 88.2-91.4\% and 89.8-92.4\% for FedProx, respectively. Similarly, for Vision Transformer, IID and non-IID scenarios see a reduction of 96.9-97.3\% and 93.9-96.8\% for FedAvg, and 96.5-97.0\% and 94.5-97.1\% for FedProx in computations. Therefore, these findings indicate that FedBaF's computational demands are relatively minimal despite the foundation model being integrated into every training round.

These findings clearly indicate that \textbf{FedBaF's computational demands are minimal}, even when integrating the foundation model into every training round, making it a highly efficient solution in both IID and non-IID environments.
\newpage
\clearpage
\section{Training curves}
\label{appendix:training_curves}
\begin{figure*}[ht]
    \centering
    \includegraphics[width=15.9cm]{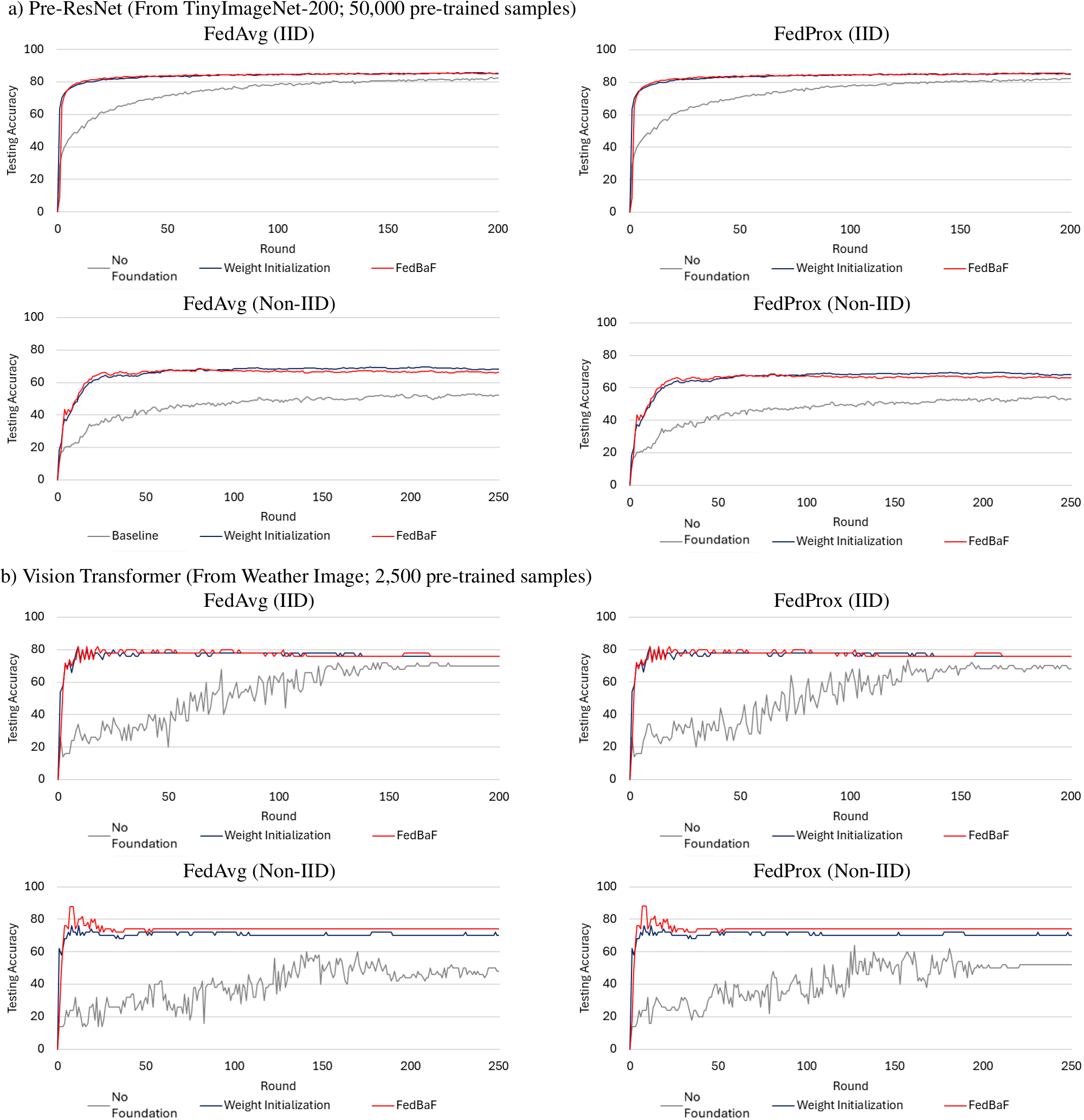}
    \vspace{-0.3 cm}
    \caption{This figure shows the evolution of model accuracy over training epochs for Pre-ResNets and Vision Transformers under IID and Non-IID scenarios. It compares the performance using no foundation model, weight initialization, and FedBaF with FedAvg and FedProx.}
    \label{figure:traning curves}
\end{figure*}

In this section, we present the evolution of model accuracy over epochs for the experiments described in Section~\ref{appendix:additional_experimental_evaluation}, as shown in Figure~\ref{figure:traning curves}. The experiments involve both Pre-ResNet and Vision Transformer models, focusing on two setups: \textbf{1)} \textbf{Pre-ResNet}, with foundation models pre-trained using TinyImageNet-200 (50,000 pre-trained samples), and \textbf{2)} \textbf{Vision Transformer}, with foundation models pre-trained using the Weather Image dataset (2,500 pre-trained samples). We present results for both IID and non-IID settings, using FedAvg and FedProx as the aggregation methods.

These training curves illustrate the progression of model accuracy throughout the training process, comparing the behaviors of models using weight initialization and FedBaF. The curves demonstrate that FedBaF performs similarly to traditional weight initialization methods and achieves higher accuracies than when no foundation model is used.

\clearpage
\newpage
\section{Convergence Analysis} \label{appendix:theory}
This section provides a convergence analysis for the FedBaF algorithm under non-convex settings using Stochastic Gradient Descent (SGD). The study demonstrates how integrating a foundation model during aggregation can improve convergence rates, even in non-IID scenarios.

\subsection{Problem Setup}
Consider the global objective function in federated learning, which is defined as:
\begin{equation}
F(\mathbf{w}) = \frac{1}{\sum_{k=1}^{m} n_k} \sum_{k=1}^{m} n_k f_{D_k}(\mathbf{w}),
\end{equation}
where \( m \) is the number of clients, $f_{D_k}(\mathbf{w})$ represents the local objective function on client $k$, $n_k$ is the number of data samples at client $k$, $\mathbf{w} \in \mathbb{R}^d$ are the model weights, \( F(\mathbf{w}) \) is the global objective function. \textit{To simplify our analysis, we assume all clients participate in every global round.}

We define $\mathbf{w}^{(t,\ell)}_k$ as the model weights of client $k$ at global round $t$ and local update step $\ell$. Therefore, before the first local update step in each global round, each client's model weights $\mathbf{w}_k^t$ are equal to the global model weights $\mathbf{w}_t$ and
\begin{equation}
    \mathbf{w}^{(t,0)}_k=\mathbf{w}_k^t=\mathbf{w}_t.
\end{equation}

\subsubsection{Assumptions}
\label{sec:appendix_assumptions}
We make the following standard assumptions to facilitate the convergence analysis:

\paragraph{L-Smoothness:} Each local objective function $f_{D_k}(\mathbf{w})$ is $L$-smooth with respect to $\mathbf{w}$, meaning:
\begin{equation}
f_{D_k}(\mathbf{v}) \leq f_{D_k}(\mathbf{w}) + \nabla f_{D_k}(\mathbf{w})^\top (\mathbf{v} - \mathbf{w}) + \frac{L}{2} \|\mathbf{v} - \mathbf{w}\|^2, \quad \forall \mathbf{w}, \mathbf{v} \in \mathbb{R}^d.
\label{eq:l_smoothness_local_epochs_1}
\end{equation}
\text{and}
\begin{equation}
\|\nabla f_{D_k}(\mathbf{v}) - \nabla f_{D_k}(\mathbf{w})\| \leq L \|\mathbf{v} - \mathbf{w}\|, \quad \forall \mathbf{w}, \mathbf{v} \in \mathbb{R}^d.
\label{eq:l_smoothness_local_epochs_2}
\end{equation}

\paragraph{Unbiased Mini-Batch Gradients:} The stochastic gradients computed over mini-batches during local updates are unbiased estimates of the true gradients.
\begin{equation}
\mathbb{E}\left[ g_{D_k}(\mathbf{w}; b)
    \right]= \nabla f_{D_k}(\mathbf{w}) \quad \forall \mathbf{w} \in \mathbb{R}^d.
    \label{eq:unbiased_gradients_local_epochs}
\end{equation}

\paragraph{Bounded Gradient Norm:} The gradients of the local objective functions are bounded by a constant \( \beta \). Specifically,
\begin{equation}
\|\nabla f_{D_k}(\mathbf{w})\|\leq \beta \quad \forall \mathbf{w} \in \mathbb{R}^d.
    \label{eq:gradient_norm_bound}
\end{equation}

\paragraph{Bounded Gradient Noise:} 
\begin{equation}
    \mathbb{E}\left[\|g_{D_k}(\mathbf{w}_k^{(t,\ell)}; b_{k,\ell})\|^2\right] \leq G^2 + B^2 \|\nabla F(\mathbf{w}_t)\|^2
\end{equation}
where \( G^2 \) and \( B^2 \) are constants that bound the dissimilarity, \( b_{k,\ell} \) denotes the \( j \)-th mini-batch on client \( k \), and \( \ell \) indexes the local update steps during global round \( t \).

\paragraph{Foundation Model Alignment:} The foundation model shares the same architecture as the global model, guaranteeing seamless integration during aggregation.

\subsection{FedBaF Aggregation Step with Multiple Local Updates}
The FedBaF algorithm operates in global rounds, where each global round $t$ includes multiple local iterations denoted by $\ell$. Each client performs multiple local updates in each global round.

The global model is updated at round $t$ using the rule:
\begin{equation}
\mathbf{w}_{t+1} = \frac{1}{1 + \alpha_t \tau_t} \left(\mathbf{w}'_t + \alpha_t \tau_t \mathbf{w}_{\text{pre}}\right),
\end{equation}
where $\mathbf{w}'_t$ is the aggregated model from the client updates, $\mathbf{w}_{\text{pre}}$ is the foundation model's weights, $\alpha_t$ is a scaling factor, and $\tau_t$ represents the correction factor based on the foundation model.

Each client $k$ performs multiple local SGD updates over local iterations $\ell = 0, 1, \dots, \Lambda_k-1$, where $\Lambda_k$ is the number of local updates for each client $k$. For each local update, the local model is updated as:
\begin{equation}
\mathbf{w}_k^{(t,\ell+1)} = \mathbf{w}_k^{(t,\ell)} - \eta g_{D_k}(\mathbf{w}_k^{(t,\ell)}; b_{k,\ell}),
\end{equation}
where $g_{D_k}(\mathbf{w}_k^{(t,\ell)}; b_{k,\ell})$ is the stochastic gradient computed on the mini-batch $b_{k,\ell}$ at local iteration $\ell$.

At the end of the local updates, each client sends the model $\mathbf{w}_k^{(t,\Lambda_k)}$ to the server, where the global model $\mathbf{w}'_t$ is computed as the weighted average of the client updates:
\begin{equation}
\mathbf{w}'_t = \mathbf{w}_t - \eta \sum_{k=1}^{m} p_k \sum_{\ell=0}^{\Lambda_k-1} g_{D_k}(\mathbf{w}_k^{(t,\ell)}; b_{k,\ell}),
\end{equation}
where $p_k = \frac{n_k}{\sum_{k=1}^{m} n_k}$ is the weight of client $k$.

Substituting this into the FedBaF update rule:
\begin{equation}
\mathbf{w}_{t+1} = \frac{1}{1 + \alpha_t \tau_t} \left(\mathbf{w}_t + \alpha_t \tau_t \mathbf{w}_{\text{pre}} - \eta \sum_{k=1}^{m} p_k\sum_{\ell=0}^{\Lambda_k-1} g_{D_k}(\mathbf{w}_k^{(t,\ell)}; b_{k,\ell})\right)
\end{equation}

For small values of $\alpha_t \tau_t \ll  1$, we use the fact that:
\begin{equation}
\begin{aligned}
    \frac{1}{1 + \alpha_t \tau_t} &= \sum_{i=0}^\infty (- \alpha_t \tau_t)^i\\
    &= 1 - \alpha_t \tau_t + \sum_{i=2}^\infty (- \alpha_t \tau_t)^i\\
    &\approx 1 - \alpha_t \tau_t
\end{aligned}
\end{equation}

which simplifies the update rule to:
\begin{equation}
\begin{aligned}
\mathbf{w}_{t+1} &\approx \mathbf{w}_t - \eta \sum_{k=1}^{m} p_k\sum_{\ell=0}^{\Lambda_k-1} g_{D_k}(\mathbf{w}_k^{(t,\ell)}; b_{k,\ell}) \\
&\hspace{20mm}- \alpha_t \tau_t \left( \left[\mathbf{w}_t-\eta \sum_{k=1}^{m} p_k\sum_{\ell=0}^{\Lambda_k-1} g_{D_k}(\mathbf{w}_k^{(t,\ell)}; b_{k,\ell}) \right] -\left( 1 - \alpha_t \tau_t \right)\mathbf{w}_{\text{pre}}\right).   
\end{aligned}
\end{equation}

The update rule can now be interpreted as multiple local gradient descent steps combined with a bias correction.
Here, we define the correction term $\chi_t$ as follows: \\
\begin{equation}
    \chi_t = \alpha_t \tau_t \left( \left[\mathbf{w}_t-\eta \sum_{k=1}^{m} p_k\sum_{\ell=0}^{\Lambda_k-1} g_{D_k}(\mathbf{w}_k^{(t,\ell)}; b_{k,\ell}) \right] -\left( 1 - \alpha_t \tau_t \right)\mathbf{w}_{\text{pre}}\right).
\end{equation}
$\chi_t$ acts as a correction that adjusts the direction of the gradient descent to leverage the foundation model's knowledge.
The update rule becomes:
\begin{equation}
    \mathbf{w}_{t+1} \approx \mathbf{w}_t - \eta \sum_{k=1}^{m} p_k\sum_{\ell=0}^{\Lambda_k-1} g_{D_k}(\mathbf{w}_k^{(t,\ell)}; b_{k,\ell}) - \chi_t.
\end{equation}

\subsection{Decrease in Objective Function}
Using the smoothness property of $ F(\mathbf{w}) $:
\begin{equation}
F(\mathbf{w}_{t+1}) \leq F(\mathbf{w}_t) + \nabla F(\mathbf{w}_t)^\top (\mathbf{w}_{t+1} - \mathbf{w}_t) + \frac{L}{2} \|\mathbf{w}_{t+1} - \mathbf{w}_t\|^2.
\end{equation}

We substitute the update rule:
\begin{equation}
\mathbf{w}_{t+1} - \mathbf{w}_t \approx -\eta \sum_{k=1}^{m} p_k \sum_{\ell=0}^{\Lambda_k-1} g_{D_k}(\mathbf{w}_k^{(t,\ell)}; b_{k,\ell}) - \chi_t.
\end{equation}

Thus:
\begin{equation}
\begin{aligned}
        F(\mathbf{w}_{t+1}) 
    \lessapprox& F(\mathbf{w}_t) - \eta \nabla F(\mathbf{w}_t)^\top \sum_{k=1}^{m} p_k \sum_{\ell=0}^{\Lambda_k-1} g_{D_k}(\mathbf{w}_k^{(t,\ell)}; b_{k,\ell}) - \nabla F(\mathbf{w}_t)^\top \chi_t \\ 
    &+ \frac{L}{2} \|\eta \sum_{k=1}^{m} p_k \sum_{\ell=0}^{\Lambda_k-1} g_{D_k}(\mathbf{w}_k^{(t,\ell)}; b_{k,\ell})- \chi_t\|^2.    
\end{aligned}
\end{equation}

Given the update rule, the change in the objective function can be bounded using the triangle inequality:
\begin{equation}
    \begin{aligned}
    F(\mathbf{w}_{t+1}) - F(\mathbf{w}_t) 
    \lessapprox& -\eta \nabla F(\mathbf{w}_t)^\top \sum_{k=1}^{m} p_k \sum_{\ell=0}^{\Lambda_k-1} g_{D_k}(\mathbf{w}_k^{(t,\ell)}; b_{k,\ell}) - \nabla F(\mathbf{w}_t)^\top \chi_t \\
    & + L\eta^2 \|\sum_{k=1}^{m} p_k \sum_{\ell=0}^{\Lambda_k-1} g_{D_k}(\mathbf{w}_k^{(t,\ell)}; b_{k,\ell})\|^2 + L\|\chi_t\|^2 \label{eq:change_in_objective}
    \end{aligned}
\end{equation}

\subsubsection{Taking Expectations}
We now take expectations of both sides of~\eqref{eq:change_in_objective}, based on the unbiasedness of mini-batch gradients and the assumptions about bounded gradient norms and smoothness.

We now compute the expectation of the change in the objective function:

\begin{equation}
    \begin{aligned}
&\mathbb{E}\left[\nabla F(\mathbf{w}_t)^\top\left(-\eta\sum_{k=1}^m p_k \sum_{\ell=0}^{\Lambda_k-1} g_{D_k}(\mathbf{w}_k^{(t,\ell)}; b_{k,\ell})\right)\right]\\
&= \nabla F(\mathbf{w}_t)^\top \left(-\eta (\sum_{k=1}^m p_k\Lambda_k) \nabla F(\mathbf{w}_t) + \eta (\sum_{k=1}^m p_k\Lambda_k) \nabla F(\mathbf{w}_t)  -\eta\sum_{k=1}^m p_k \sum_{\ell=0}^{\Lambda_k-1} \mathbb{E}\left[g_{D_k}(\mathbf{w}_k^{(t,\ell)}; b_{k,\ell})\right]\right)\\
&= -\eta (\sum_{k=1}^m p_k\Lambda_k)\|\nabla F(\mathbf{w}_t)\|^2 + \nabla F(\mathbf{w}_t)^\top \left(\eta (\sum_{k=1}^m p_k\Lambda_k) \sum_{k=1}^m p_k \nabla f_{D_k}(\mathbf{w}_t) -\eta\sum_{k=1}^m p_k \sum_{\ell=0}^{\Lambda_k-1} \nabla f_{D_k}(\mathbf{w}_k^{(t,\ell)})\right).
    \end{aligned}
\end{equation}
Next, we simplify the remaining term:
\begin{equation}
    \begin{aligned}
&= -\eta (\sum_{k=1}^m p_k\Lambda_k) \|\nabla F(\mathbf{w}_t)\|^2 + \eta \sum_{k=1}^m p_k  \sum_{\ell=0}^{\Lambda_k-1}  \nabla F(\mathbf{w}_t)^\top \left( \frac{\sum_{k=1}^m p_k\Lambda_k}{\Lambda_k} \nabla f_{D_k}(\mathbf{w}_t) - \nabla f_{D_k}(\mathbf{w}_k^{(t,\ell)})\right) \\
&\leq -\eta (\sum_{k=1}^m p_k\Lambda_k)\|\nabla F(\mathbf{w}_t)\|^2 + \frac{\eta}{2} \sum_{k=1}^m p_k  \sum_{\ell=0}^{\Lambda_k-1}  \left[\|\nabla F(\mathbf{w}_t)\|^2 + \|\frac{\sum_{k=1}^m p_k\Lambda_k}{\Lambda_k}\nabla f_{D_k}(\mathbf{w}_t) - \nabla f_{D_k}(\mathbf{w}_k^{(t,\ell)})\|^2\right].
    \end{aligned}
\end{equation}
Finally, using the fact that the gradient is bounded, we get:
\begin{equation}
    \begin{aligned}
&\leq -\left(\frac{\eta}{2}\sum_{k=1}^m p_k\Lambda_k\right) \|\nabla F(\mathbf{w}_t)\|^2 + \frac{\eta}{2}\sum_{k=1}^m p_k  \sum_{\ell=0}^{\Lambda_k-1} \|\frac{\sum_{k=1}^m p_k\Lambda_k}{\Lambda_k}\nabla f_{D_k}(\mathbf{w}_t) - \nabla f_{D_k}(\mathbf{w}_k^{(t,\ell)})\|^2\\
&\leq -\left(\frac{\eta}{2}\sum_{k=1}^m p_k\Lambda_k\right) \|\nabla F(\mathbf{w}_t)\|^2 + \frac{\eta}{2}\sum_{k=1}^m 2p_k  \sum_{\ell=0}^{\Lambda_k-1} \left[\|  \frac{\sum_{k=1}^m p_k\Lambda_k}{\Lambda_k}\nabla f_{D_k}(\mathbf{w}_t)\|^2 + \|\nabla f_{D_k}(\mathbf{w}_k^{(t,\ell)})\|^2\right]\\
&\leq -\left(\frac{\eta}{2}\sum_{k=1}^m p_k\Lambda_k\right) \|\nabla F(\mathbf{w}_t)\|^2 + \eta\beta^2 \sum_{k=1}^m p_k \left(\frac{(\sum_{k=1}^m p_k\Lambda_k)^2}{\Lambda_k} + \Lambda_k \right).
    \end{aligned}
\end{equation}

This provides a bound on the expected decrease in the global objective function.

Similarly,
\begin{equation}
    \begin{aligned}
    &\mathbb{E}\left[L\eta^2 \|\sum_{k=1}^{m} p_k \sum_{\ell=0}^{\Lambda_k-1} g_{D_k}(\mathbf{w}_k^{(t,\ell)}; b_{k,\ell})\|^2\right]\\
    &\leq Lm\eta^2 \sum_{k=1}^{m} \mathbb{E}\left[\|p_k \sum_{\ell=0}^{\Lambda_k-1} g_{D_k}(\mathbf{w}_k^{(t,\ell)}; b_{k,\ell})\|^2\right]\\
    &\leq Lm\eta^2 \sum_{k=1}^{m} p_k^2  \Lambda_k \sum_{\ell=0}^{\Lambda_k-1} \mathbb{E}\left[\|g_{D_k}(\mathbf{w}_k^{(t,\ell)}; b_{k,\ell})\|^2\right]\\
    &\leq Lm\eta^2 \sum_{k=1}^{m} p_k^2  \Lambda_k^2 (G^2 + B^2 \|\nabla F(\mathbf{w}_t)\|^2)
    \end{aligned}
\end{equation}

Plugging these bounds into~\eqref{eq:change_in_objective} gives
\begin{equation}
    \begin{aligned}
\mathbb{E}&\left[F(\mathbf{w}_{t+1}) - F(\mathbf{w}_t) \right]
\lessapprox -\eta \nabla F(\mathbf{w}_t)^\top \sum_{k=1}^{m} p_k \sum_{\ell=0}^{\Lambda_k-1} g_{D_k}(\mathbf{w}_k^{(t,\ell)}; b_{k,\ell}) - \nabla F(\mathbf{w}_t)^\top \mathbb{E}\left[\chi_t\right] \\
&\hspace{20mm} + L\eta^2\|\sum_{k=1}^{m} p_k \sum_{\ell=0}^{\Lambda_k-1} g_{D_k}(\mathbf{w}_k^{(t,\ell)}; b_{k,\ell})\|^2 + L\mathbb{E}\left[\|\chi_t\|^2\right] \\
&\leq -\left(\frac{\eta}{2}\sum_{k=1}^m p_k\Lambda_k\right) \|\nabla F(\mathbf{w}_t)\|^2 + \eta\beta^2 \sum_{k=1}^m p_k \left(\frac{(\sum_{k=1}^m p_k\Lambda_k)^2}{\Lambda_k} + \Lambda_k \right) - \nabla F(\mathbf{w}_t)^\top \mathbb{E}\left[\chi_t\right] \\
&\hspace{20mm} + Lm\eta^2 \sum_{k=1}^{m} p_k^2  \Lambda_k^2 (G^2 + B^2 \|\nabla F(\mathbf{w}_t)\|^2) + L\mathbb{E}\left[\|\chi_t\|^2\right] \\
&\leq \left(-\frac{\eta}{2}\sum_{k=1}^m p_k\Lambda_k + B^2Lm\eta^2 \sum_{k=1}^{m} p_k^2  \Lambda_k^2\right) \|\nabla F(\mathbf{w}_t)\|^2 + \eta\beta^2 \sum_{k=1}^m p_k \left(\frac{(\sum_{k=1}^m p_k\Lambda_k)^2}{\Lambda_k} + \Lambda_k \right) \\
&\hspace{20mm}- \nabla F(\mathbf{w}_t)^\top \mathbb{E}\left[\chi_t\right]+ Lm\eta^2 \sum_{k=1}^{m} p_k^2  \Lambda_k^2 G^2 + L\mathbb{E}\left[\|\chi_t\|^2\right]
    \end{aligned}
\end{equation}

\subsection{Summing Over Iterations}
To establish a convergence result, we sum this inequality over $ T $ iterations:
\begin{equation}
\begin{aligned}
\sum_{t=1}^{T} F(\mathbf{w}_{t+1}) \leq &\sum_{t=1}^{T} F(\mathbf{w}_t) + \left(-\frac{\eta}{2}\sum_{k=1}^m p_k\Lambda_k + B^2Lm\eta^2 \sum_{k=1}^{m} p_k^2  \Lambda_k^2\right) \sum_{t=1}^{T}\|\nabla F(\mathbf{w}_t)\|^2\\
&+ \eta\beta^2 T \sum_{k=1}^m p_k \left(\frac{(\sum_{k=1}^m p_k\Lambda_k)^2}{\Lambda_k} + \Lambda_k \right) - \sum_{t=1}^{T}\nabla F(\mathbf{w}_t)^\top \mathbb{E}\left[\chi_t\right]  \\
&+ Lm\eta^2 T \sum_{k=1}^{m} p_k^2  \Lambda_k^2 G^2 
+ L\sum_{t=1}^{T}\mathbb{E}\left[\|\chi_t\|^2\right] 
\end{aligned}
\end{equation}

As long as $\eta < \frac{\sum_{k=1}^m p_k\Lambda_k}{2B^2Lm \sum_{k=1}^{m} p_k^2  \Lambda_k^2}$, the term $\frac{1}{2}\sum_{k=1}^m p_k\Lambda_k - B^2Lm\eta \sum_{k=1}^{m} p_k^2  \Lambda_k^2$ is positive.
Therefore, rearranging and dividing by $ T $ gives:
\begin{equation}
\begin{aligned}
    \frac{1}{T} \sum_{t=1}^{T} \mathbb{E}[\|\nabla F(\mathbf{w}_t)\|^2] \leq& \frac{F(\mathbf{w}_1) - F(\mathbf{w}_{T+1})}{T(\frac{1}{2}\sum_{k=1}^m p_k\Lambda_k - B^2Lm\eta \sum_{k=1}^{m} p_k^2  \Lambda_k^2)} \\
    &+ \frac{\eta\beta^2 \sum_{k=1}^m p_k \left(\frac{(\sum_{k=1}^m p_k\Lambda_k)^2}{\Lambda_k}+ \Lambda_k \right) + Lm\eta^2 \sum_{k=1}^{m} p_k^2  \Lambda_k^2 G^2}
    {\frac{1}{2}\sum_{k=1}^m p_k\Lambda_k - B^2Lm\eta \sum_{k=1}^{m} p_k^2  \Lambda_k^2}\\
    & +\frac{L\sum_{t=1}^{T}\mathbb{E}\left[\|\chi_t\|^2\right] - \sum_{t=1}^{T}\nabla F(\mathbf{w}_t)^\top \mathbb{E}\left[\chi_t\right] }
    {T(\frac{1}{2}\sum_{k=1}^m p_k\Lambda_k - B^2Lm\eta \sum_{k=1}^{m} p_k^2  \Lambda_k^2)} 
    \end{aligned}
\end{equation}

If the sign of 
\[
L\mathbb{E}[\|\chi_t\|^2] - \nabla F(\mathbf{w}_t)^\top\mathbb{E}[\chi_t] 
\]
is negative, we obtain a tighter bound, which implies that the correction terms positively influence convergence.

We know that:
\[
\chi_t = \alpha_t \tau_t \left( \left[ \mathbf{w}_t - \eta \sum_{k=1}^{m} p_k \sum_{\ell=0}^{\Lambda_k-1} g_{D_k}(\mathbf{w}_k^{(t,\ell)}; b_{k,\ell}) \right] - \left( 1 - \alpha_t \tau_t \right) \mathbf{w}_{\text{pre}}  \right).
\]

This means:
\[
\mathbb{E}\left[\|\chi_t\|^2\right] = \alpha_t^2 \tau_t^2 \left\| \left[ \mathbf{w}_t - \eta \sum_{k=1}^{m} p_k \sum_{\ell=0}^{\Lambda_k-1}  \nabla f_{D_k}(\mathbf{w}_k^{(t,\ell)}) \right] - \left( 1 - \alpha_t \tau_t \right) \mathbf{w}_{\text{pre}} \right\|^2.
\]

Given that \( \alpha_t \tau_t \) is small enough at a sufficiently large global round 
$t$, the higher-order terms with \( \alpha_t^2 \tau_t^2\) become negligible. Therefore, for large $t$, $[\|\chi_t\|^2] \approx 0$. Next, the direction of the correction $\chi_t$ and the direction of the gradient of the global objective $\nabla F(\mathbf{w}_t)$ agreeing implies that the
inner product $\nabla F(\mathbf{w}_t)^\top \mathbb{E}\left[\chi_t\right]$ is positive.

We conclude that:
\[
L \mathbb{E}\left[\|\chi_t\|^2\right] < \nabla F(\mathbf{w}_t)^\top \mathbb{E}[\chi_t].
\]

This shows that the variance of the correction term \( \chi_t \) is significantly smaller than its impact on the inner product, leading to a tighter convergence bound, especially when \( \alpha_t \tau_t \) is small but positive.

\subsection{Influence of the Correction Term}
The correction term \( \chi_t \), derived from the foundation model, plays a significant role in the convergence behavior. The influence of \( \chi_t \) ensures that the gradient descent step is adjusted based on the foundation model's knowledge. By controlling the size of \( \alpha_t \tau_t \), the foundation model can guide the global model towards better solutions, especially in non-IID scenarios. The correction term provides additional stability and enhances convergence, particularly when the local models exhibit significant heterogeneity.

\clearpage
\newpage
\section{Proofs of Propositions} \label{appendix:proposition}
\subsection{Proof of Proposition \ref{prop:improvement}} 
\improvementprop*
\begin{proof}
We present a convergence analysis of our FL framework that incorporates foundation models in the aggregation phase according to Alg.~\ref{algorithm: adapting pre-trained model} \textit{Lines 8-10}.
By comparing the square distance between $\mathbf{w}_{t+1}$ and $\mathbf{w}^*$ to the square distance between $\mathbf{w}'_{t+1}$ and $\mathbf{w}^*$,
we derive conditions under which our method converges to $\mathbf{w}^*$ faster than FedAvg. Noting that $\forall t\;\alpha_t,\tau_t \geq 0$,
\begin{align}
    \|\mathbf{w}_{t+1} - \mathbf{w}^*\|^2 &=  \|\frac{1}{1 + \alpha_t\tau_t} (\mathbf{w}'_{t+1} + \alpha_t\tau_t \mathbf{w}_{\text{pre}}) - \mathbf{w}^*\|^2 \nonumber\\
    &=  \frac{1}{(1 + \alpha_t\tau_t)^2}\|(\mathbf{w}'_{t+1} -\mathbf{w}^*) +  \alpha_t\tau_t (\mathbf{w}_{\text{pre}} -\mathbf{w}^*)\|^2 \nonumber\\
    &\leq  \frac{\|\mathbf{w}'_{t+1} -\mathbf{w}^*\|^2 +  \alpha_t^2\tau_t^2\|\mathbf{w}_{\text{pre}} -\mathbf{w}^*\|^2}{1 + 2\alpha_t\tau_t + \alpha_t^2\tau_t^2} \label{eq:squared_bound}
\end{align}

For notational convenience, we define $\beta_t := \|\mathbf{w}'_{t+1} -\mathbf{w}^*\|$ and $\gamma := \|\mathbf{w}_{\text{pre}} -\mathbf{w}^*\|$.
FedBaF is better than FedAvg when the right side is less than $\beta_t^2$. So, we upper bound the right side by $\beta_t^2$ and find values of $\alpha_t$ that satisfy the bound.
{
\begin{align*}
    \frac{\beta_t^2 + \alpha_t^2\tau_t^2\gamma^2}{1 + 2\alpha_t\tau_t + \alpha_t^2\tau_t^2} &< \beta_t^2\\
    \beta_t^2 + \alpha_t^2\tau_t^2\gamma^2 &< \beta_t^2 + 2\alpha_t\tau_t\beta_t^2 + \alpha_t^2\tau_t^2\beta_t^2\\
    \alpha_t^2\tau_t^2(\gamma^2-\beta_t^2) - 2\alpha_t\tau_t\beta_t^2 &< 0
\end{align*} 
}
Note that $\alpha_t$ is sampled from the uniform distribution $\frac{\psi}{\tau_0}\text{U}(1,2)$. For the above inequality to be satisfied for a given $t$, there are three cases: 
\begin{enumerate}
    \item $\beta_t>\gamma$: This case occurs when $t$ is small and $\mathbf{w}^*$ is closer to $\mathbf{w}_{\text{pre}}$ than $\mathbf{w}^*$ is to the FedBaF global model. In this case, we require
    $$\alpha_t > \frac{2\beta_t^2}{(\gamma^2-\beta_t^2)\tau_t}$$
    $\alpha_t$ always satisfies this inequality since the RHS is negative and $\alpha_t > 0$ by definition.
    \item $\beta_t=\gamma$: This means that we require $\alpha_t > 0$, which is always true by definition.
    \item $\beta_t<\gamma$: This case may occur when $t$ is large and $\mathbf{w}^*$ is closer to the FedBaF global model than $\mathbf{w}^*$ is to $\mathbf{w}_{\text{pre}}$. In this case, we get a meaningful bound for $\alpha_t$:
    $$\alpha_t < \frac{2\beta_t^2}{(\gamma^2-\beta_t^2)\tau_t}$$

\end{enumerate}

When $\forall t\;\alpha_t$ satisfies the above conditions, the proposition holds.
\end{proof}

\clearpage
\newpage
\subsection{Proof of Proposition \ref{prop:divergence}} 
\divergenceprop*
\begin{proof}
\begin{align*}
   \|\mathbf{w}_{t} - \mathbf{w}^*\| &= \left\|\frac{1}{1 + \alpha_t\tau_t} (\mathbf{w}'_{t} + \alpha_t\tau_t \mathbf{w}_{\text{pre}}) - \mathbf{w}^*\right\| \\
   &= \frac{\left\| (\mathbf{w}'_{t} - \mathbf{w}^*) + \alpha_t\tau_t(\mathbf{w}_{\text{pre}} - \mathbf{w}^*)\right\|}{1 + \alpha_t\tau_t}\\
   &\leq \frac{ \|\mathbf{w}'_{t} - \mathbf{w}^*\| + \alpha_t\tau_t \|\mathbf{w}_{\text{pre}} - \mathbf{w}^*\|}{1 + \alpha_t\tau_t} \\
   &= \frac{\left\|\sum_{k \in S_t} \frac{n_k}{n} (\mathbf{w}_{t}^k - \mathbf{w}^*)\right\| + \alpha_t\tau_t \|\mathbf{w}_{\text{pre}} - \mathbf{w}^*\|}{1 + \alpha_t\tau_t} \\
   &\leq \frac{\sum_{k \in S_t} \frac{n_k}{n} \|\mathbf{w}_{t}^k - \mathbf{w}^*\| + \alpha_t\tau_t\|\mathbf{w}_{\text{pre}} - \mathbf{w}^*\|}{1 + \alpha_t\tau_t}\\
   &\leq\frac{\delta_t + \alpha_t\tau_t\gamma}{1 + \alpha_t\tau_t}
\end{align*}
where we set $\gamma = \|\mathbf{w}_{\text{pre}} -\mathbf{w}^*\|$.
Since non-IID data can cause significant variance in local updates, we compare the derived bound to FedAvg, where the bound on the distance between $\mathbf{w}_{t}$ and $\mathbf{w}^*$ is \(\delta_t\).
By assumption, $\gamma \leq \delta_t$ for earlier rounds (small $t$). We get \( \frac{\delta_t + \alpha_t\tau_t \gamma}{1 + \alpha_t\tau_t} \leq \delta_t \), which is equivalent to
{
\[
\delta_t + \alpha_t\tau_t \gamma \leq \delta_t + \alpha_t\tau_t\delta_t  = \delta_t(1 + \alpha_t\tau_t) 
\]
}
Therefore, 
{
\[
\|\mathbf{w}_{t} - \mathbf{w}^*\| \leq \frac{\delta_t + \alpha_t\tau_t \gamma}{1 + \alpha_t\tau_t}
\]
}
Since $\frac{\delta_t + \alpha_t\tau_t \gamma}{1 + \alpha_t\tau_t} \leq \delta_t$, FedBaF has a tighter upper bound on $\|\mathbf{w}_{t} - \mathbf{w}^*\|$ than FedAvg. This demonstrates the advantage of using a foundation model in non-IID settings.
    
\end{proof}
\newpage
\clearpage
\section{Security Analysis in the Presence of Adversarial Attacks}\label{appendix:security}
In this section, we discuss the potential for extracting a foundation model in FedBaF and demonstrate FedBaF's robustness against backdoor attacks. These attacks pose unique security challenges to FL systems, involving malicious alterations within model updates to degrade system performance or embed hidden vulnerabilities. We will analyze how FedBaF mitigates these threats and ensures integrity and security.

\subsection{Possibility of Extracting a Foundation Model}
As discussed in Section~\ref{sec: methodology}, using a randomized $\alpha_t$ prevents the extraction of the foundation model's weights. However, the aggregated global models might still exhibit components of the foundation model by following a similar weight distribution. To investigate this, we analyze the distance between the global model and the foundation model over the first 200 aggregation rounds.

Let \( \mathbf{w}_{t+1} \) and \( \mathbf{w}_{pre} \) represent the weights of the global model and the foundation model, respectively. For each weight tensor \( \mathbf{w}_{t+1}^i \) and \( \mathbf{w}_{pre}^i \) with matching shapes, we calculate the normalized distance for each element and then average these distances. For each element \( j \) in the weight tensor \(\mathbf{w}_{t+1}^i \) and \( \mathbf{w}_{pre}^i \):
\[
\text{dist}_j^i = \frac{| w_{t+1,j}^i - w_{pre,j}^i |}{|w_{t+1,j}^i|}
\]
where \( w_{t+1,j}^i \) is the \( j \)-th element of the \( i \)-th weight tensor of the global model; \( w_{pre,j}^i \) is the \( j \)-th element of the \( i \)-th weight tensor of the foundation model; and \( \text{dist}_j^i \) is the normalized distance for the \( j \)-th element of the \( i \)-th weight tensor.

We concatenate all element-wise distances \( \text{dist}_j^i \) across all weight tensors and then compute the mean of these distances:
\[
\text{Dist} = \frac{1}{N_{param}} \sum_{i=1} \sum_{j=1} \text{dist}_j^i
\]
where \( N_{param} \) is the total number of elements across all matching weight tensors and \( \text{Dist} \) is the overall average normalized distance.

\begin{figure}[ht]
    \centering
    \includegraphics[width=\linewidth]{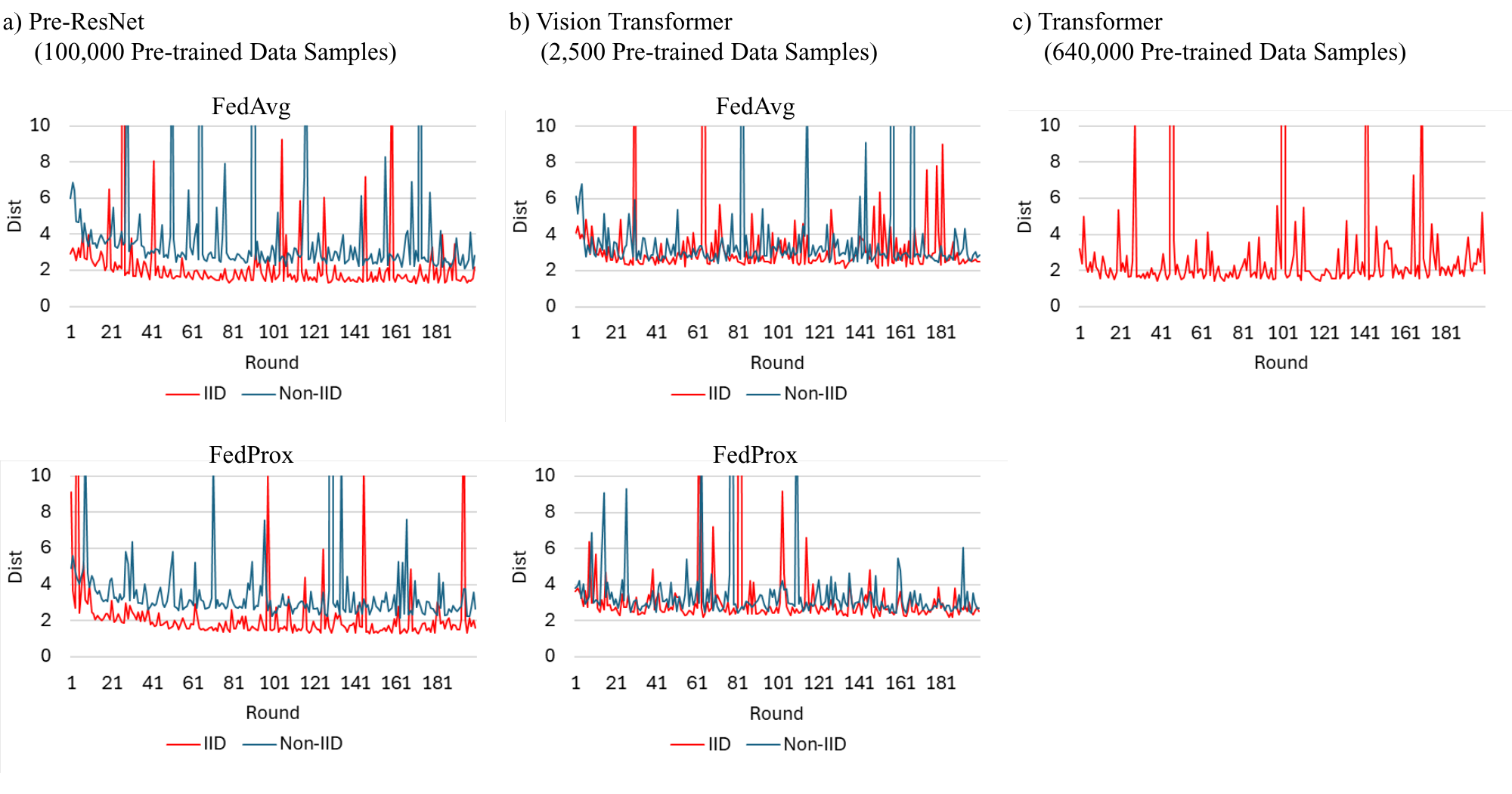}
    \vspace{-0.5 cm}
    \caption{Distances (\(\text{Dist}\)) between the global model and the foundation model across aggregation rounds. The minimum \(\text{Dist}\) observed was 1.27, indicating significant differences in scale between the foundation model's weights and the aggregated global model's weights.}
    \label{fig: distance_two_models}
\end{figure}

In Figure~\ref{fig: distance_two_models}, the curves show the distances, \(\text{Dist}\), for each aggregation round. The minimum \(\text{Dist}\) across all cases was 1.27, indicating that the distance has a 127\% scale of the magnitude of the weights of the aggregated global model. This means the foundation model's weights differ in scale from the aggregated weights. To analyze the effect of distance intensity, we added Gaussian random noise based on the magnitude of each foundation model's weights to the foundation model's weights.

\begin{figure}[ht]
    \centering
    \includegraphics[width=\linewidth]{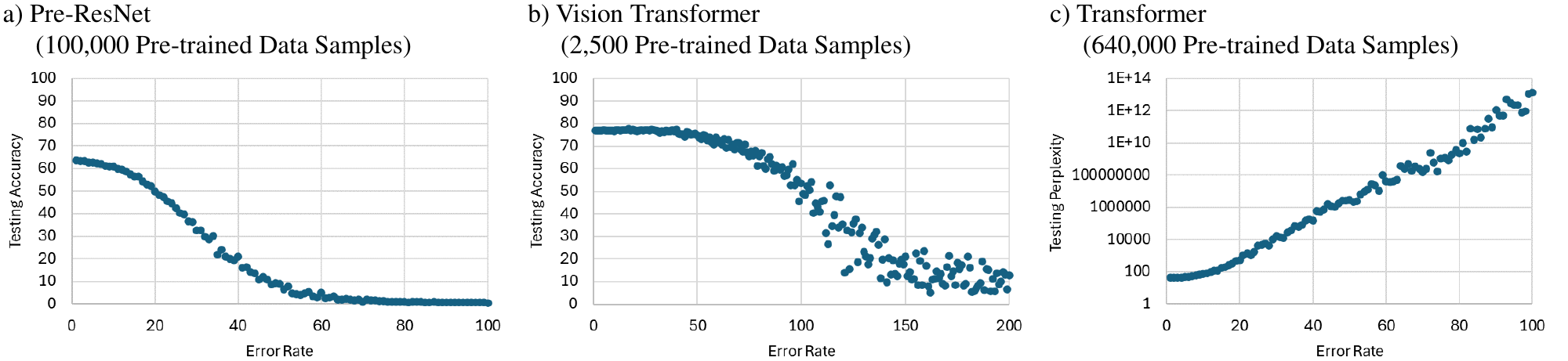}
    \vspace{-0.5 cm}
    \caption{Testing accuracy according to the added noise. The x-axis error rate is calculated by the magnitude of the added Gaussian noise divided by the magnitude of the foundation model's weights.}
    \label{fig: adding noise}
\end{figure}

Figure~\ref{fig: adding noise} shows the testing accuracy as a function of the added noise. The x-axis represents the error rate, calculated by dividing the magnitude of the added Gaussian noise by the magnitude of the foundation model's weights. We used the best-performing foundation models from those with varying pre-trained sample sizes, as described in Section~\ref{appendix:additional_experimental_evaluation}. When a 127\% error rate is applied, the Pre-ResNet model shows almost 0\% testing accuracy, the Vision Transformer shows 30\% testing accuracy, and the Transformer model exhibits excessively high testing perplexity. This empirical evidence indicates that extracting the foundation model's knowledge is impossible after training begins from the global model. The diverse updates during training in FedBaF significantly disrupt the alignment between the foundation model's weights and the global model, preventing any meaningful extraction of the foundation model's information.

To this end, we examine the proximity of the global model, $\mathbf{w}_t$, to the foundation model and to the averaged local models, $\mathbf{w}_t'$, throughout the training process. We first determine the distances between $\mathbf{w}_t'$ and $\mathbf{w}_t$ and between $\mathbf{w}_t$ and $\mathbf{w}_{pre}$:
\begin{align*}
\|\mathbf{w}_t - \mathbf{w}_t'\| &= \left\|\frac{1}{1 + {\alpha}_t\tau_{t}} (\mathbf{w}_t' + {\alpha}_t\tau_{t}\mathbf{w}_{pre}) - \mathbf{w}_t'\right\|\\
& = \frac{{\alpha}_t\tau_{t}}{1 + {\alpha}_t\tau_{t}} \|\mathbf{w}_t' - \mathbf{w}_{pre}\| \\
\|\mathbf{w}_t - \mathbf{w}_{pre}\| &= \left\|\frac{1}{1 + {\alpha}_t\tau_{t}} \mathbf{w}_t' + {\alpha}_t\tau_{t}\mathbf{w}_{pre} - \mathbf{w}_{pre}\right\|\\
& = \frac{1}{1 + {\alpha}_t\tau_{t}} \|\mathbf{w}_t' - \mathbf{w}_{pre}\|
\end{align*}
At the onset of training, both distances are equivalent since we make a strategic choice for the weight $\alpha_t\tau_0$ to be approximately 2. This simplifies the initial update rule for the global model such that the initial global model weights, $\mathbf{w}_0$, are an unweighted average of the client's updated model weights $\mathbf{w}_0'$ and the foundation model weights $\mathbf{w}_{pre}$.
As the training progresses, ${\alpha}_t\tau_{t}$ typically decays to less than 1. We deduce for $t>0$:
\begin{align*}
    \frac{{\alpha}_t\tau_{t}}{1 + {\alpha}_t\tau_{t}} < \frac{1}{1 + {\alpha}_t\tau_{t}} \implies \|\mathbf{w}_t - \mathbf{w}_t'\| < \|\mathbf{w}_t - \mathbf{w}_{pre}\|
\end{align*}

As $t\rightarrow\infty$, $\mathbf{w}_t$ will drift away from $\mathbf{w}_{pre}$ and towards $\mathbf{w}_t'$.
Due to the intricate dissemination of learned insights across all model weights, and the complexities of high-dimensional weight spaces, it is difficult to reverse-engineer $\mathbf{w}_{pre}$ from $\mathbf{w}_t$. Even a subset of weights does not provide enough information to predict the rest deterministically. The inherent complexity of the model weight (parameter) space is a natural defense mechanism in FedBaF.

\subsection{Mitigating Backdoor Attacks}
\textbf{Backdoor attacks} in FL involve embedding a dormant malicious function in a local model. Integrating foundation models mitigates such attacks by diluting the impact of individual client updates. Specifically, we have the updates
\begin{align*}
\Delta \mathbf{w}_{\text{client}}^t &= \text{ClientUpdate}(\mathbf{w}_t) \\
\mathbf{w}_{t+1} &= \frac{1}{1 + \alpha_t\tau_t} (\Delta \mathbf{w}_{\text{client}}^t + \alpha_t\tau_t \mathbf{w}_{\text{pre}})
\end{align*}
Here, \( \Delta \mathbf{w}_{\text{client}}^t \) is the update from client \( c \) at iteration \( t \), and \( \mathbf{w}_{\text{pre}} \) is the foundation model weight. The factor \( \tau_t \) controls the influence of the foundation model.
This mathematical formulation showcases the security benefits of our method. By incorporating the foundation model, the aggregation counterbalances the (malicious) client update \( \Delta \mathbf{w}_{\text{client}}^t \).
This approach thus \textit{enhances the system's resilience to adversarial attacks} by maintaining a consistent learning direction and reducing the impact of compromised updates.

\subsection{Experiments on Variations of \texorpdfstring{$\tau_t$}.}
\begin{figure}[ht]
    \centering
    \includegraphics[width=\linewidth]{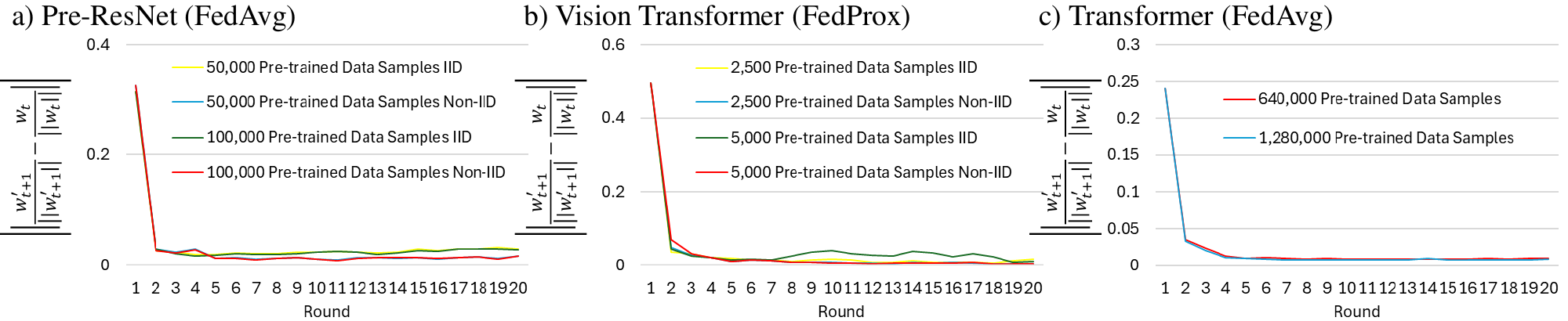}
    \vspace{-0.5 cm}
    \caption{Variations of $\|\frac{\mathbf{w}'_{t+1}}{\|\mathbf{w}'_{t+1}\|} - \frac{\mathbf{w}_{t}}{\|\mathbf{w}_{t}\|}\|\;(=\tau_t\sqrt{t+1})$ across training rounds.}
    \label{fig: changes of taus}
\end{figure}
Figure~\ref{fig: changes of taus} illustrates the non-adversarial IID and non-IID scenarios from Tables~\ref{tab:test result resnet}, \ref{tab:test result vit}, and \ref{tab:test result transformer}. We observed that the numerator of $\tau_t$ (as referenced in Alg.~\ref{algorithm: adapting pre-trained model} \textit{Line 9}) consistently decreases towards 0, independent of the denominator ($\sqrt{t+1}$), after several rounds. This indicates that FedBaF benefits from the foundation model's guidance but retains the ability to effectively and quickly adapt to new data.

\end{document}